\documentclass[journal]{IEEEtran}
\usepackage{cite}
\usepackage{amssymb}
\usepackage{hyperref}
\usepackage{graphicx}
\usepackage{subfigure}
\usepackage[table,xcdraw]{xcolor}
\usepackage{multirow}
\usepackage{overpic}
\usepackage{amsmath}

\ifCLASSINFOpdf
\else
\fi
\begin{document}
%

\title{Spatio-Temporal Self-Attention Network\\ for Video Saliency Prediction}
%
%
%

\author{Ziqiang~Wang,
        Zhi~Liu,~\IEEEmembership{Senior Member,~IEEE,}
        Gongyang~Li,~\IEEEmembership{Member,~IEEE,}
        Yang~Wang,\\
        Tianhong~Zhang,
        Lihua~Xu,
        and~Jijun~Wang
\thanks{This work was supported in part by the National Natural Science Foundation of 
China under Grants 62171269 and 82171544, in part by the Science and Technology 
Commission of Shanghai Municipality under Grant 21S31903100,
and in part by the China Scholarship Council under Grant 202006890079. 
\textit{(Corresponding author: Zhi Liu.)}}
\thanks{Ziqiang Wang, Zhi Liu, and Gongyang Li are with Shanghai Institute for 
Advanced Communication and Data Science, Shanghai University, 
Shanghai 200444, China, and also with the School of Communication 
and Information Engineering, Shanghai University, Shanghai 200444, China.
Gongyang Li is also with the School of Computer Science and Engineering, 
Nanyang Technological University, Singapore 639798 
(e-mail: wziqiang@shu.edu.cn; liuzhisjtu@163.com; ligongyang@shu.edu.cn).}
\thanks{Yang Wang is with the Department of Computer Science, 
University of Manitoba, Winnipeg, MB R3T 2N2, Canada 
(e-mail: ywang@cs.umanitoba.ca).}
\thanks{Tianhong Zhang, Lihua Xu, and Jijun Wang are with Shanghai Key Laboratory 
of Psychotic Disorders, Shanghai Mental Health Center, 
Shanghai Jiaotong University School of Medicine, Shanghai 200030, China
(e-mail: zhang\_tianhong@126.com; dr\_xulihua@163.com; dr\_wangjijun@126.com).}
}

%
%

\markboth{IEEE TRANSACTIONS ON MULTIMEDIA}%
{Shell \MakeLowercase{\textit{et al.}}: Bare Demo of IEEEtran.cls for IEEE Journals}
%

\maketitle
\begin{abstract}
3D convolutional neural networks have achieved promising results for video tasks in computer vision, 
including video saliency prediction that is explored in this paper.
However, 3D convolution encodes visual representation merely on fixed local spacetime according to its kernel size, while human attention is always attracted by relational visual features at different time.
To overcome this limitation, we propose a novel Spatio-Temporal Self-Attention 3D Network (STSANet) for video saliency prediction,
in which multiple Spatio-Temporal Self-Attention (STSA) modules are employed at different levels of 3D convolutional backbone 
to directly capture long-range relations between spatio-temporal features of different time steps.
Besides, we propose an Attentional Multi-Scale Fusion (AMSF) module to integrate multi-level features 
with the perception of context in semantic and spatio-temporal subspaces.
Extensive experiments demonstrate the contributions of key components of our method,
and the results on DHF1K, Hollywood-2, UCF, and DIEM benchmark datasets clearly prove the superiority of the proposed model
compared with all state-of-the-art models.

\end{abstract}


%
\IEEEpeerreviewmaketitle

\section{Introduction}
%
%
%
%

\begin{figure}[t]
  \flushright
  \small
  \begin{overpic}[width=.47\textwidth]{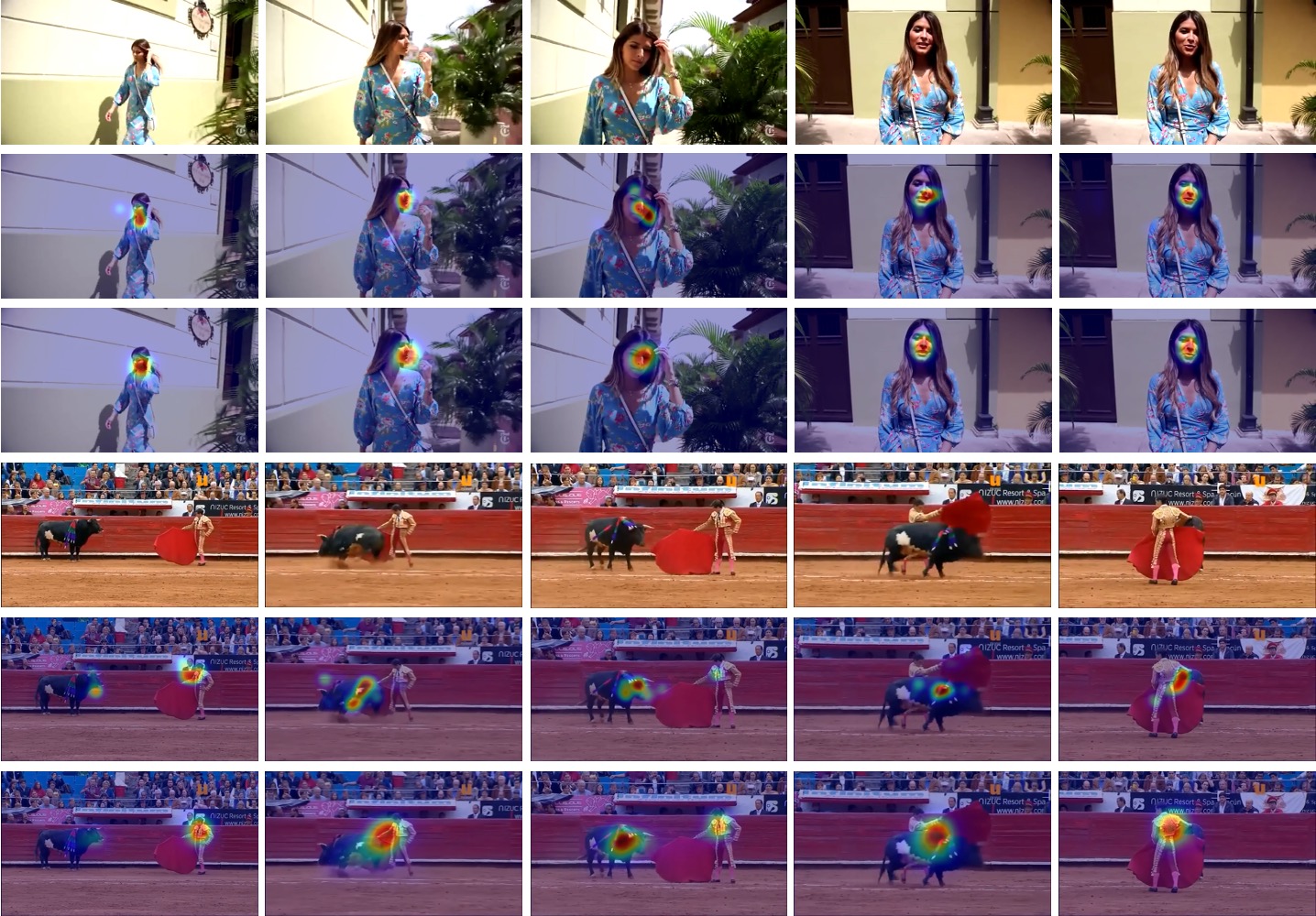}
    \put(-3,4){\rotatebox{90}{SP}}
    \put(-3,14){\rotatebox{90}{GT}}
    \put(-3,24){\rotatebox{90}{Frame}}
    \put(-3,39){\rotatebox{90}{SP}}
    \put(-3,50){\rotatebox{90}{GT}}
    \put(-3,59){\rotatebox{90}{Frame}}
  \end{overpic}
  \caption{
      Visualization of results of the proposed STSANet on sampling frames in two videos.
  }
  \label{Fig:visualization}
\end{figure}
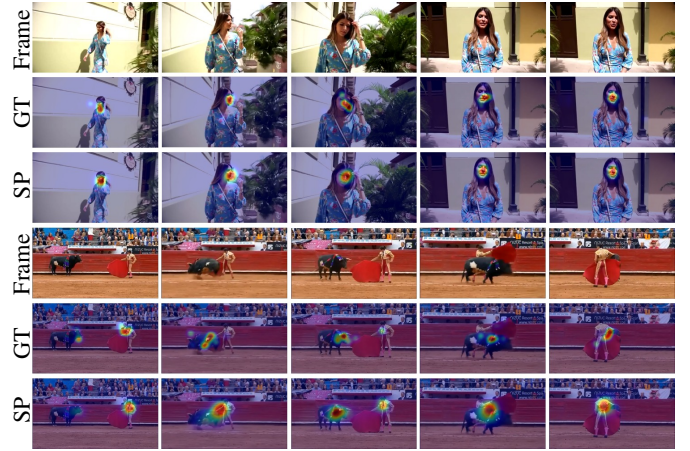

 
\IEEEPARstart{H}{umans} have a fantastic capability of localizing the most important area in the visual field promptly, 
named as visual attention mechanism, which facilitates the processing of diverse visual information.
In computer vision, modeling the visual attention mechanism is a fundamental research topic, 
named \emph{saliency prediction} (SP) or \emph{fixation prediction},
which aims to deduce the visual saliency degree of each region in images, 
expressed in the form of a saliency map (as shown in \autoref{Fig:visualization}).
SP has been widely applied to various computer vision tasks, such as image captioning \cite{zhou2020captioning}, photo cropping \cite{wang2019cropping},
object segmentation \cite{li2019personal,li2021personal, wang2018VOS, wang2021VOS}, video compression \cite{hadizadeh2014compression}, \textit{etc}.

Traditional solutions for SP leverage hand-crafted features, including low-level cues such as  color, orientation, and intensity,
as well as high-level contents such as persons and objects \cite{itti1998, le2006bottom-up, cerf2008face}.
In the recent years, with the renaissance of neural networks and advance of high-quality datasets and benchmarks \cite{MIT1003, SALICON_dataset},
deep learning based SP has obtained substantial progress,
and the computational models have performed close to human inter-observer level on static datasets.
Compared with image SP, video saliency prediction (VSP) is more difficult and develops relatively slowly.

Videos contain spatial information in frames and temporal information between frames.
In videos, human attention is not only guided by low-level cues and semantic context in a single frame, 
but also by relations of features in frames.
For example, in a video clip, the same object moving in a scene usually attracts visual attention, as shown in \autoref{Fig:visualization}.
Consequently, it is crucial for video saliency prediction (VSP) to synchronously exploit spatial and temporal information.
Directly using image SP models for VSP triggers poor performance,
due to ignoring the temporal information.
Recently, deep learning models for VSP have emerged and outperformed the traditional models \cite{mahadevan2010spatiotemporal, fang2014video}.
Some VSP models \cite{bak2018two-stream, wu2019fusion, zhang2019two-stream} employ both RGB and optical flow backbones to encode appearance and motion information, respectively, 
and merge them for dynamic saliency inference.
However, the motion stream merely considers temporal relations between adjacent frames.
This limitation is alleviated by LSTM-based models \cite{jiang2018DeepVS, wang2021ACLNet, lai2020STRA-Net, chen2021BiLSTM, zhang2021Recurrent, wu2020SalSAC}, 
which adopt LSTMs to capture temporal long-term relations in a video.
These models utilize convolutional networks and LSTMs to deal with spatial and temporal information separately, 
therefore, they are unable to synchronously exploit spatial and temporal information, which is instrumental in VSP.
To this end, some models~\cite{min2019TASED-Net, bellitto2020HD2S, jain2020ViNet} use 3D convolutional networks to
process spatio-temporal information jointly, and have shown progressive performance.

In this paper, we further explore VSP based on 3D convolutional networks.
Although 3D convolution can encode spatio-temporal information collectively, 
it processes fixed local spacetime and fails to capture long-range relations between visual features at different time.
To remedy this deficiency, we propose a novel Spatio-Temporal Self-Attention (STSA) module 
to directly learn long-range spatio-temporal dependencies, which is inspired by the self-attention mechanism \cite{vaswani2017attention}.
In the STSA module, spatio-temporal features are split along the temporal channel, 
and features at different time steps are separately transformed to an embedding space by embedding convolutional layers.
Afterwards, long-range spatio-temporal interactions are built by dot-product attention calculation between features at different time steps.

Based on the STSA module, we propose a Spatio-Temporal Self-Attention Network (STSANet) for VSP.
The backbone of our model is a 3D fully convolutional network from S3D \cite{xie2018S3D} pre-trained on Kinetics dataset \cite{2017kinetics}.
We draw four branches from different levels of the backbone, and employ STSA modules on them, respectively,
to produce local and global spatio-temporal context at multiple levels.
Since the spatio-temporal and semantic gaps exist in the outputs from the four STSA modules,
direct fusion, such as summation and concatenation, is not powerful enough to process them well.
Instead, we design an Attentional Multi-Scale Fusion (AMSF) module to integrate the multi-level features.
The AMSF module consists of an attentional weighting operation and a spatio-temporal multi-scale structure,
which are used to alleviate the semantic and spatio-temporal gaps, respectively, during feature fusion.

Overall, our main contributions can be summarized as follows:
\begin{itemize}
\item 
We propose a Spatio-Temporal Self-Attention Network (STSANet) for video saliency prediction,
in which self-attention mechanism is integrated into the 3D convolutional network.
This integration helps our model achieve superior performance compared with all state-of-the-art models on multiple benchmark datasets.

\item 
We propose a Spatio-Temporal Self-Attention (STSA) module to capture long-range dependencies between time steps of temporal channel.
The STSA module is employed at multiple levels of 3D convolutional network to complement the locality of 3D convolution
and to produce local and global spatio-temporal representations for video saliency prediction.

\item 
We propose an Attention Multi-Scale Fusion (AMSF) module to fuse spatio-temporal features from STSA modules arranged at different levels of backbone.
The AMSF module perceives contextual contents in semantic and spatio-temporal subspaces, and narrows semantic and spatio-temporal gaps during saliency feature fusion.
\end{itemize}

\section{Related Work}
In computer vision, saliency models can be divided into two types: 
\emph{saliency prediction} (SP) and \emph{salient object detection} (SOD).
SP aims to predict visual saliency degree of each region in images,
while the task objective of SOD is highlighting salient object regions.
Numerous computational models have been proposed for image SOD \cite{zhang2010SOD, wang2019SOD, zhou2019RGB, wang2020SOD, ma2020SOD, ren2021SOD, wang2021SOD_survey} 
and video SOD \cite{liu2017videoSOD, zhou2018VSOD, wang2018VSOD, fan2019VSOD} in recent decades.
Also, SP contains image SP and video SP, and this work focuses on video SP (VSP).
In this section, we briefly review the SP models and the network design related to our work.

\subsection{Image Saliency Prediction}
The earliest work for SP, proposed by Itti \textit{et al.} \cite{itti1998}, captured hand-crafted futures in color, intensity, and orientation channels,
respectively, and combined them for saliency results.
After this work, SP models based on hand-crafted features emerged consecutively \cite{harel2007Graph-based, le2006bottom-up, cerf2008face}.
Recently, with the advance of deep learning, data-driven models have outperformed traditional models.
\emph{Ensembles of Deep Networks} (eDN) \cite{vig2014eDN} is among the first to apply neural networks to SP task.
This model combined features generated from a lot of convolutional layers using a linear classifier.
\emph{Deep Gaze \uppercase\expandafter{\romannumeral1}}~\cite{2015DeepGaze1} employed the off-the-shelf features from deep convolutional neural network (CNN) trained on ImageNet \cite{deng2009imagenet}
for SP, and \emph{Salicon}~\cite{2015salicon} further fine-tuned pre-trained VGGNet~\cite{VGGNet} with SP datasets, 
which verified the effectiveness of transfer learning for SP.
After that, a variety of deep SP models~\cite{2015salicon, 2016SalNet, 2016PDP, 2016ML-Net, 2017DeepFix, 2017DeepGaze2, 2018DSCLRCN, 2018MxSalNet, 2018DVA, 2018SAM, 2020DINet, 2020GazeGAN, 2020MSI-Net, 2020SimpleNet}
based on VGGNet \cite{VGGNet}, ResNet \cite{ResNet}, DenseNet \cite{DenseNet}, 
and NASNet \cite{NASNet} have been proposed successively.
At present, SP models have performed close to human inter-observer level on image SP datasets.

\subsection{Video Saliency Prediction}
Traditional models related to VSP mainly explored dynamic scene viewing using static and motion information \cite{mahadevan2010spatiotemporal, fang2014video},
however, hand-crafted spatio-temporal features were not powerful enough for modeling dynamic saliency.
A number of VSP models based on deep learning have emerged in recent years.
\subsubsection{Two-Stream Methods}
Bak \textit{et al.} \cite{bak2018two-stream} proposed a two-stream network, which employed two convolutional backbones, with RGB images and optical flow maps as inputs respectively,
to extract spatio-temporal information and fuse their outputs for saliency inference.
Wu \textit{et al.}~\cite{wu2019fusion} and Zhang \textit{et al.}~\cite{zhang2019two-stream} further explored the two-stream structure 
and investigated fusion methods to improve performance.
\subsubsection{LSTM-based Methods}
However, the optical flow network merely considers temporal relations between adjacent frames, 
hence LSTM is frequently adopted to extend temporal perception.
Gorji \textit{et al.} \cite{gorji2018push} employed multi-stream LSTMs that merged with static SP model for VSP.
DeepVS \cite{jiang2018DeepVS} leveraged object and motion networks to extract intra-frame saliency information, 
and modeled temporal correlation between frames by convLSTMs.
Besides, ACLNet \cite{wang2021ACLNet} designed a neural attention module in a CNN-LSTM structure, which was supervised with image SP datasets.
STRA-Net \cite{lai2020STRA-Net} employed dense residual cross connection to enrich interactions between motion stream and appearance stream during feature extraction,
and incorporated an attention mechanism to enhance the spatio-temporal information.
SalEMA \cite{linardos2019SalEMA} modified the static SP model by adding a simple exponential moving average for feature fusion in temporal domain, resulting in a low-parametric architecture.
Later, Zhang \textit{et al.} \cite{zhang2021Recurrent} utilized spatial and channel attention to select and re-weight spatio-temporal information, 
and employed an attentive convLSTM to model relations between frames.
SalSAC \cite{wu2020SalSAC} designed a correlation-based convLSTM for VSP, 
in which adjacent frames were weighted according to similarity between them.
\subsubsection{3D Convolution-based Methods}
Differently, TASED-Net \cite{min2019TASED-Net} proposed a 3D fully-convolutional encoder-decoder network for VSP and achieved promising performance.
Compared with LSTM-based architectures, which process spatial and temporal information separately, 
a 3D network encodes and decodes spatio-temporal information in a collective way.
Moreover, ViNet \cite{jain2020ViNet} designed a UNet-like encoder-decoder network based on a 3D backbone, 
in which features from multiple levels were upsampled with trilinear upsampling and concatenated along the temporal channel.
In HD2S \cite{bellitto2020HD2S}, multi-level features from a 3D encoder are separately decoded to obtain single-channel conspicuity maps,
and integrated them for saliency results.
Besides, TSFP-Net \cite{chang2021TSFP} employs a feature pyramid structure with top-down feature integration on a 3D convolutional backbone,
and combines the multi-level spatio-temporal features to reason the saliency result for a video frame.
In our work, we further explore 3D neural network for VSP.
The 3D convolution handles a local spatio-temporal space at a time, 
therefore, 3D convolutional networks lack the ability to directly model long-range spatio-temporal relations.
Our model addresses this shortage by employing STSA modules, 
which build long-range immediate interactions between spatio-temporal features of different time steps,
at multiple levels of the 3D backbone.

\subsection{Self-Attention}
Self-attention mechanism has been an important issue after \cite{vaswani2017attention}, 
in which a lot of self-attention mechanisms were employed to learn text representations for natural language tasks.
In self-attention mechanism, input tokens are linearly transformed as queries, keys, and values, respectively, in embedding layers, 
and then long-range relations between tokens of input sequence are calculated by dot-product attention.

In computer vision, Wang \textit{et al.} \cite{wang2018non-local} proposed the non-local neural network 
that is a classical implementation of self-attention mechanism in vision tasks.
Besides, Oh \textit{et al.} \cite{oh2019STM} designed memory networks based on self-attention mechanism for video object segmentation.
In that model, the memory was updated according to the dependencies 
between current frame and past frames computed in the form of self-attention mechanism.
Moreover, Wang \textit{et al.} \cite{wang2021TMA} adopted the memory and self-attention mechanism for video semantic segmentation.
These two models both used 2D deep convolutional networks to encode the frames one by one,
and then transformed features of different frames to an embedding space by embedding layers for attention calculation.
Therefore, the amount of computer memory occupied is considerable.

Analogously, COSNet \cite{wang2019co-att} performs a co-attention mechanism \cite{2016coatt, 2017coatt} for video object segmentation task.
This model uses 2D convolutional networks to separately generate embeddings from current frame and a group of reference frames which are uniformly sampled from a video. 
Subsequently, the co-attention part computes attention summaries from current frame and each reference frame in pairs to capture inter-frame consistency, 
and then integrates the average of these attention summaries to the embedding of current frame to enhance the capacity of inferring target object.

In our VSP model, we adopt a 3D convolutional network as backbone
to directly extract spatio-temporal information from multiple frames,
and employ self-attention mechanism to capture long-range dependencies between spatio-temporal features of different time steps.
Meanwhile, we further adopt other strategies to compress our model as described in \autoref{Sec:STSA module}.

\begin{figure}[t]  
  \centering
  \includegraphics[width=.48\textwidth]{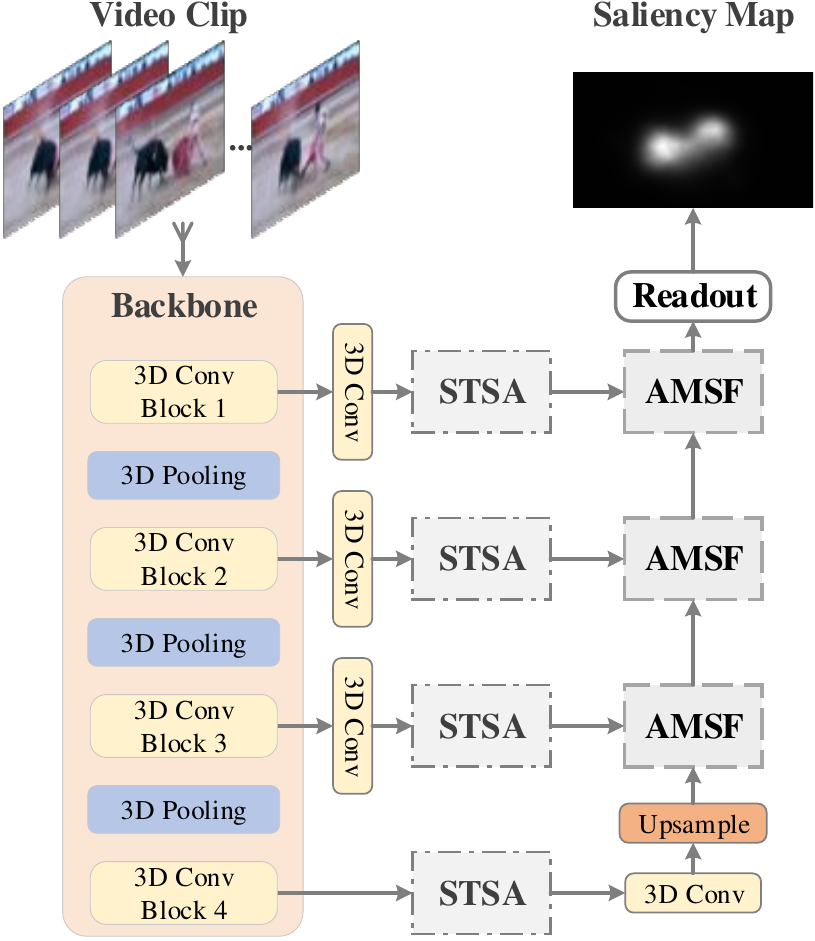}
  \caption{
      An overview of the proposed model. 
      Our model contains three main components: 3D convolutional encoder, Spatio-Temporal Self-Attention (STSA) module, and Attentional Multi-Scale Fusion (AMSF) module.
      A video clip including consecutive frames is used as the input of 3D encoder to generate hierarchical spatio-temporal visual features.
      Four STSA modules are employed after different 3D convolutional blocks to capture long-range spatio-temporal dependencies between time steps at multiple levels.
      Afterwards, multi-level features are fused by the AMSF modules to generate video saliency results.
  }
  \label{Fig:overview}
\end{figure}
\begin{figure*}[t]
  \centering
  \subfigure[]{
  \label{Fig:STSA} 
  \includegraphics[width=7.1in]{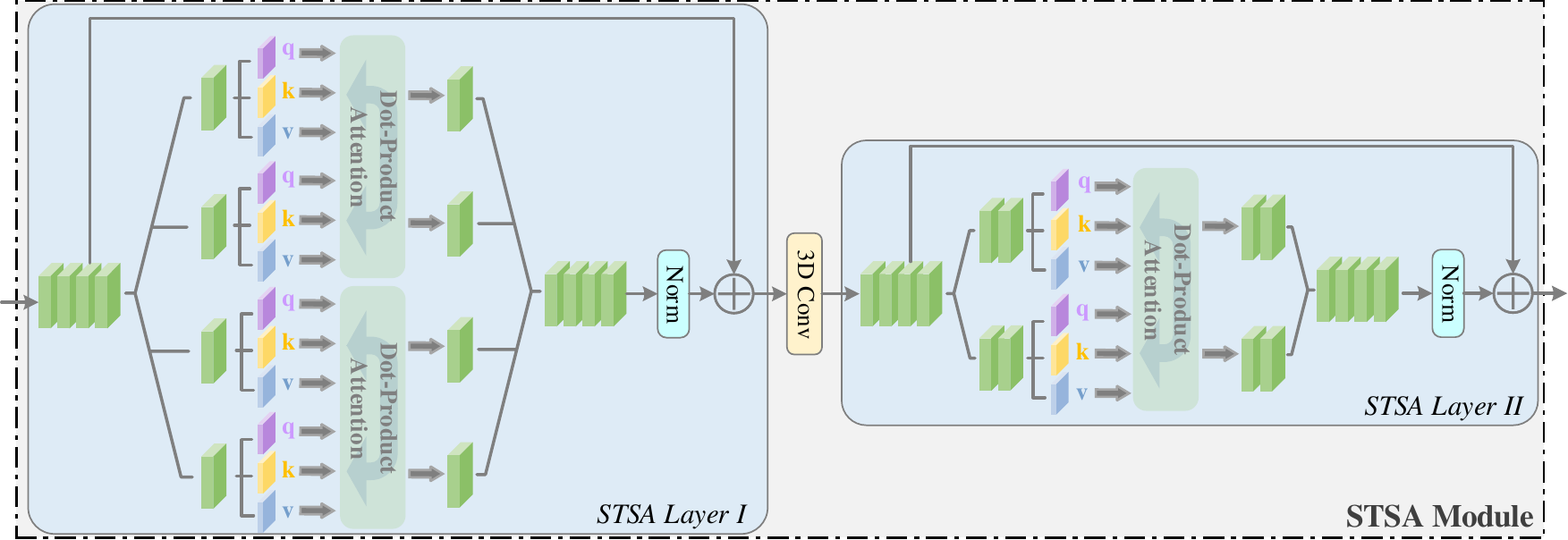}}
  \subfigure[]{
  \label{Fig:attention} 
  \includegraphics[width=4.8in]{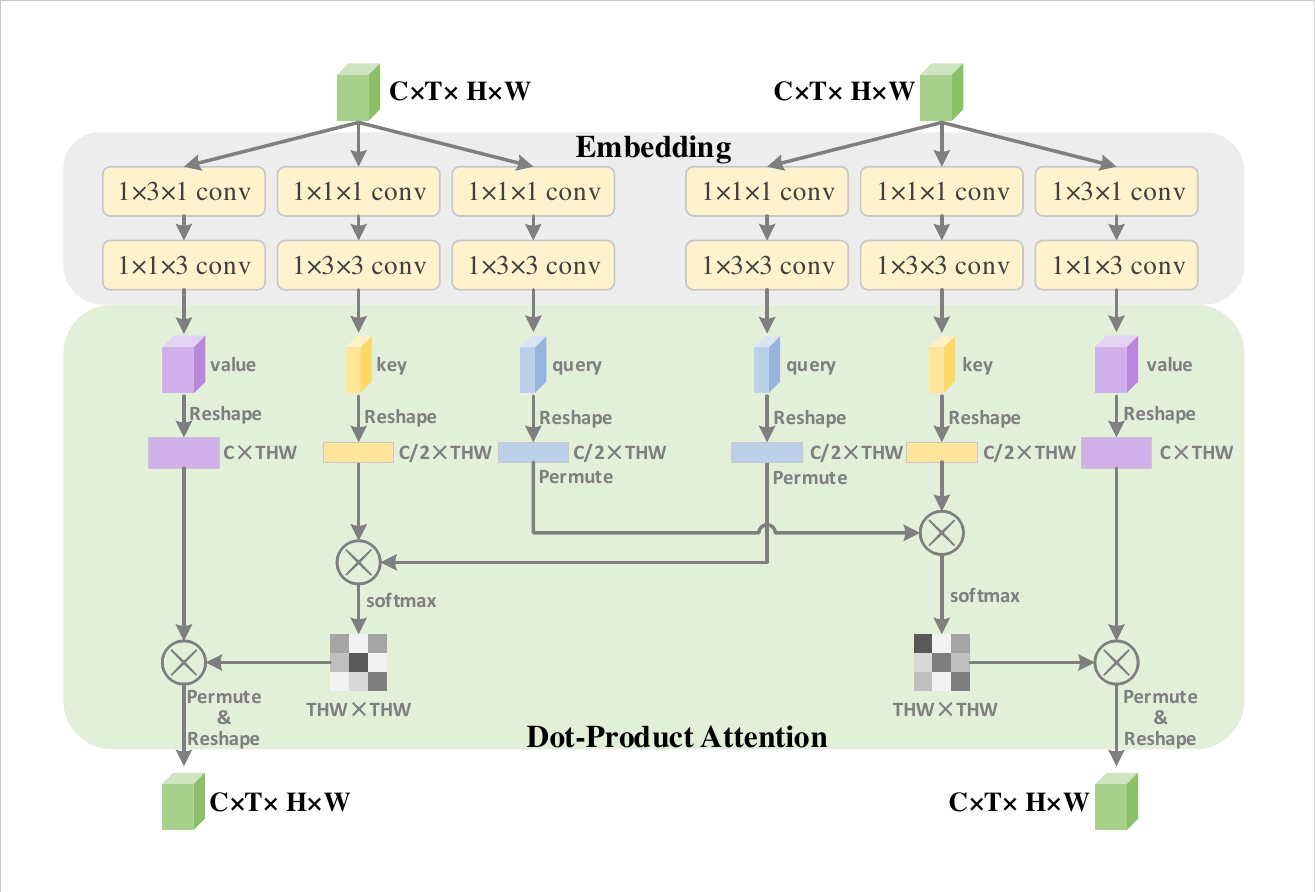}}
  \hspace{0.1in}
  \subfigure[]{
  \label{Fig:bottleneck} 
  \includegraphics[width=2.0in]{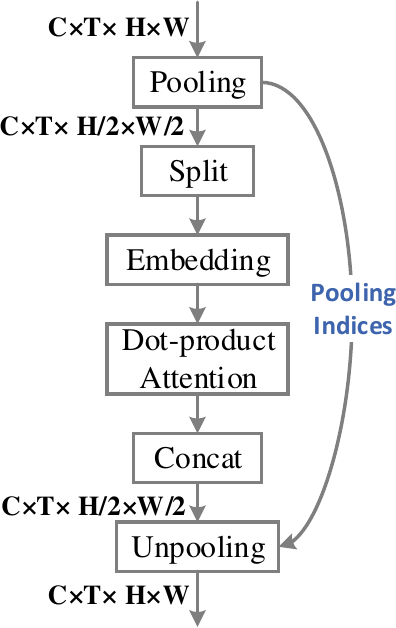}}
  \caption{
    (a) Spatio-Temporal Self-Attention (STSA) module.
    (b) An detailed illustration of the embedding and dot-product attention in STSA layers.
    (c) Spatial bottleneck structure employed on the STSA layers after $Conv\_Block\_1$.
  }
\end{figure*}
\section{The Proposed Model}
In this section, we illustrate the proposed Spatio-Temporal Self-Attention Network (STSANet).
In \autoref{Sec:overview}, an overview of our model is given.
In \autoref{Sec:STSA module}, we describe the proposed Spatio-Temporal Self-Attention (STSA) module in detail.
In \autoref{Sec:AMSF module}, we introduce the Attentional Multi-Scale Fusion module.
In \autoref{Sec:implementation details}, we provide the detailed information of the modules.
In \autoref{Sec:supervision and loss function}, we present the supervision manner and the loss function.

\subsection{Architecture Overview}
\label{Sec:overview}
The architecture of the proposed model is described in \autoref{Fig:overview}.
We utilize the fully-convolutional portion of S3D network \cite{xie2018S3D} pre-trained on the Kinetics dataset\cite{2017kinetics} as the backbone.
The backbone is composed of 3D convolutional layers, 
which have the capability of encoding spatio-temporal information.
The input of backbone is a video clip consisting of $T$ consecutive frames, set to $32$ in our model.

CNNs are able to encode hierarchical representations, 
including low-level cues like color contrast, and high-level semantic information like persons or objects,
all of which are of value to saliency.
Analogously, from shallow to deep layers in 3D CNNs, 
the outputs correspond to low- and high-level features, respectively.
Accordingly, we employ four decoding branches for the output from four 3D convolutional blocks of the backbone,
which are separated by three 3D max pooling layers.
Each branch has a Spatio-Temporal Self-Attention (STSA) module to directly create global spatio-temporal
context at each level.
After that, the features from four branches are integrated by Attentional Multi-Scale Fusion (AMSF) modules in a top-down pathway.
Lastly, in the readout module, the feature maps are upsampled to the original video resolution,
and the features at all time steps are aggregated by 3D convolution to obtain the final saliency map.

\subsection{Spatio-Temporal Self-Attention Module}
\label{Sec:STSA module}
Convolutional operator updates a position of a feature map by aggregating information in a local window,
thereby failing to capture long-range relations between visual features at different time,
which play an important role in VSP.
To complement the locality of convolutional operator, we propose a Spatio-Temporal Self-Attention (STSA) module
that further updates each position at each time step by aggregating global relations with spatio-temporal features at the other time steps.

In \autoref{Fig:overview}, after the backbone, 
the temporal channel dimensions of multi-level features are uniformly compressed to $4$ by 3D convolutional layers.
Subsequently, the input features to STSA modules can be denoted as $\mathbf{F} \in \mathbb{R}^{C \times 4 \times H \times W}$,
where $C$, $4$, $H$, and $W$ are the dimensions of the semantic channel, temporal channel, height and width, respectively.
As described in \autoref{Fig:STSA}, the STSA module cascades two STSA layers and a 3D convolutional layer in the middle.
In the \emph{STSA layer \uppercase\expandafter{\romannumeral1}}, the input feature $\mathbf{F}$ is first split to four sub-features 
$\{\mathbf{F}^{1}, \mathbf{F}^{2}, \mathbf{F}^{3}, \mathbf{F}^{4}\} \in \mathbb{R}^{C \times 1 \times H \times W}$
along the temporal channel, and each represents spatio-temporal information at different time of a video clip.
Then, we use convolutional embedding layers to transform the sub-features to queries, keys, and values, 
expressed as $\mathbf{F}^{t}_q \in \mathbb{R}^{C/2 \times 1 \times H \times W}, \mathbf{F}^{t}_k \in \mathbb{R}^{C/2 \times 1 \times H \times W}$, 
and $\mathbf{F}^{t}_v \in \mathbb{R}^{C \times 1 \times H \times W}, t=1,2,3,4$, respectively.
After that, we implement the dot-product attention between sub-features at different time steps
to directly capture long-range relations between spatio-temporal features of different time steps.
Specifically, this attention mechanism in \emph{STSA layer \uppercase\expandafter{\romannumeral1}} can be formulated as follows:
\begin{align}
  \mathrm{DA}(\mathbf{F}^{2}_q, \mathbf{F}^{1}_k, \mathbf{F}^{1}_v) = \mathit{Softmax}({(\mathbf{F}^{2}_q)}^\mathrm{T}\mathbf{F}^{1}_k){(\mathbf{F}^{1}_v)}^\mathrm{T} \label{eq1}, \\
  \mathrm{DA}(\mathbf{F}^{1}_q, \mathbf{F}^{2}_k, \mathbf{F}^{2}_v) = \mathit{Softmax}({(\mathbf{F}^{1}_q)}^\mathrm{T}\mathbf{F}^{2}_k){(\mathbf{F}^{2}_v)}^\mathrm{T} \label{eq2}, \\
  \mathrm{DA}(\mathbf{F}^{4}_q, \mathbf{F}^{3}_k, \mathbf{F}^{3}_v) = \mathit{Softmax}({(\mathbf{F}^{4}_q)}^\mathrm{T}\mathbf{F}^{3}_k){(\mathbf{F}^{3}_v)}^\mathrm{T} \label{eq3}, \\
  \mathrm{DA}(\mathbf{F}^{3}_q, \mathbf{F}^{4}_k, \mathbf{F}^{4}_v) = \mathit{Softmax}({(\mathbf{F}^{3}_q)}^\mathrm{T}\mathbf{F}^{4}_k){(\mathbf{F}^{4}_v)}^\mathrm{T}, \label{eq4}
\end{align}
where $\mathrm{DA}(\cdot)$ is the dot-product attention calculation as depicted in \autoref{Fig:attention}, 
and $\mathit{Softmax}(\cdot)$ indicates the softmax activation function.

The queries, keys and values after embedding layers are first reshaped to $C/2 \times THW$ or $C \times THW$, $T = 1$.
Afterwards, the dot-product attention calculation is carried out between $\mathbf{F}^{1}$ and $\mathbf{F}^{2}$, $\mathbf{F}^{3}$ and $\mathbf{F}^{4}$, respectively.
More concretely, as shown in \autoref{eq1}, the similarity matrix is obtained by matrix multiplication of $\mathbf{F}^{2}_q$ 
and $\mathbf{F}^{1}_k$, and normalized by the softmax function, to represent the spatio-temporal relation between $\mathbf{F}^{2}$ and $\mathbf{F}^{1}$.
After that, the relation is integrated to $\mathbf{F}^{1}$ by multiplying the similarity matrix and $\mathbf{F}^{1}_v$.
In the same way, the value of $\mathbf{F}^{2}$ is enhanced by the dot-product attention operation as described in \autoref{eq2}.
Meanwhile, the dot-product attention calculation between $\mathbf{F}^{3}$ and $\mathbf{F}^{4}$, as formulated in \autoref{eq3} and \autoref{eq4}, 
is the same as that between $\mathbf{F}^{1}$ and $\mathbf{F}^{2}$.
After the two sets of attention calculations, the four outputs are reassembled along the temporal channel 
after being restored to $C \times 4 \times H \times W$, and an identity shortcut is added as follows:
\begin{align}
   \nonumber \hat{\mathbf{F}} =& \mathbf{F} \oplus \mathrm{Norm}([\mathrm{DA}(\mathbf{F}^{2}_q, \mathbf{F}^{1}_k, \mathbf{F}^{1}_v),\mathrm{DA}(\mathbf{F}^{1}_q, \mathbf{F}^{2}_k, \mathbf{F}^{2}_v),\\
  &  \mathrm{DA}(\mathbf{F}^{4}_q, \mathbf{F}^{3}_k, \mathbf{F}^{3}_v),\mathrm{DA}(\mathbf{F}^{3}_q, \mathbf{F}^{4}_k, \mathbf{F}^{4}_v]))
\end{align}
where $\oplus$ denotes the element-wise summation, $\mathrm{Norm}(\cdot)$ denotes the layer normalization \cite{layernorm}, 
and $[\cdot,\cdot]$ donates the concatenation.

Our elaborate design on these two STSA layers achieves that each of the four features at different time steps
($\mathbf{F}^{1}$, $\mathbf{F}^{2}$, $\mathbf{F}^{3}$ and $\mathbf{F}^{4}$)
is directly updated by long-range spatio-temporal relations with the other three through only three groups of non-overlapping dot-product attention calculations.
For instance, $\mathbf{F}^{1}$ is updated by long-range relation with $\mathbf{F}^{2}$ in aforementioned \emph{STSA layer \uppercase\expandafter{\romannumeral1}}, 
and updated by long-range relations with $\mathbf{F}^{3}$ and $\mathbf{F}^{4}$ in \emph{STSA layer \uppercase\expandafter{\romannumeral2}}.
Meanwhile, $\mathbf{F}^{2}$, $\mathbf{F}^{3}$ and $\mathbf{F}^{4}$ are updated similarly.
Specifically, after \emph{STSA layer \uppercase\expandafter{\romannumeral1}} detailed above,
\emph{STSA layer \uppercase\expandafter{\romannumeral2}} splits its input feature to two sub-features $\{\mathbf{F}^{12}$, $\mathbf{F}^{34}\} \in \mathbb{R}^{C \times 2 \times H \times W}$ along the temporal channel,
and then separately transforms them to queries, keys, and values by embedding layers.  
Subsequently, the dot-product attention can be expressed in the following formulas:
\begin{align}
  & \mathrm{DA}(\mathbf{F}^{34}_q, \mathbf{F}^{12}_k, \mathbf{F}^{12}_v) = \mathit{Softmax}({(\mathbf{F}^{34}_q)}^\mathrm{T}\mathbf{F}^{12}_k){(\mathbf{F}^{12}_v)}^\mathrm{T}, \label{eq5} \\
  & \mathrm{DA}(\mathbf{F}^{12}_q, \mathbf{F}^{34}_k, \mathbf{F}^{34}_v) = \mathit{Softmax}({(\mathbf{F}^{12}_q)}^\mathrm{T}\mathbf{F}^{34}_k){(\mathbf{F}^{34}_v)}^\mathrm{T}. \label{eq6} 
\end{align}
And the identity shortcut is added as follows:
\begin{equation}
  \hat{\mathbf{F}} = \mathbf{F} \oplus \mathrm{Norm}([\mathrm{DA}(\mathbf{F}^{34}_q, \mathbf{F}^{12}_k, \mathbf{F}^{12}_v),\mathrm{DA}(\mathbf{F}^{12}_q, \mathbf{F}^{34}_k, \mathbf{F}^{34}_v])).
\end{equation}

Notably, we employ a convolutional layer between the two STSA layers to capture more local representations.
The convolutional layer halves the dimension of feature in the semantic channel, which is the input
feature of the second STSA layer, to suppress memory consumption.
Consequently, our STSA module captures local and global features alternately for complementarity.

The STSA modules described above are employed on four branches from the backbone.
They enhance the relational visual features between different time steps at multiple levels, 
which are instrumental in localizing dynamic saliency.

\begin{figure}[t]  
  \centering
  \includegraphics[width=.49\textwidth]{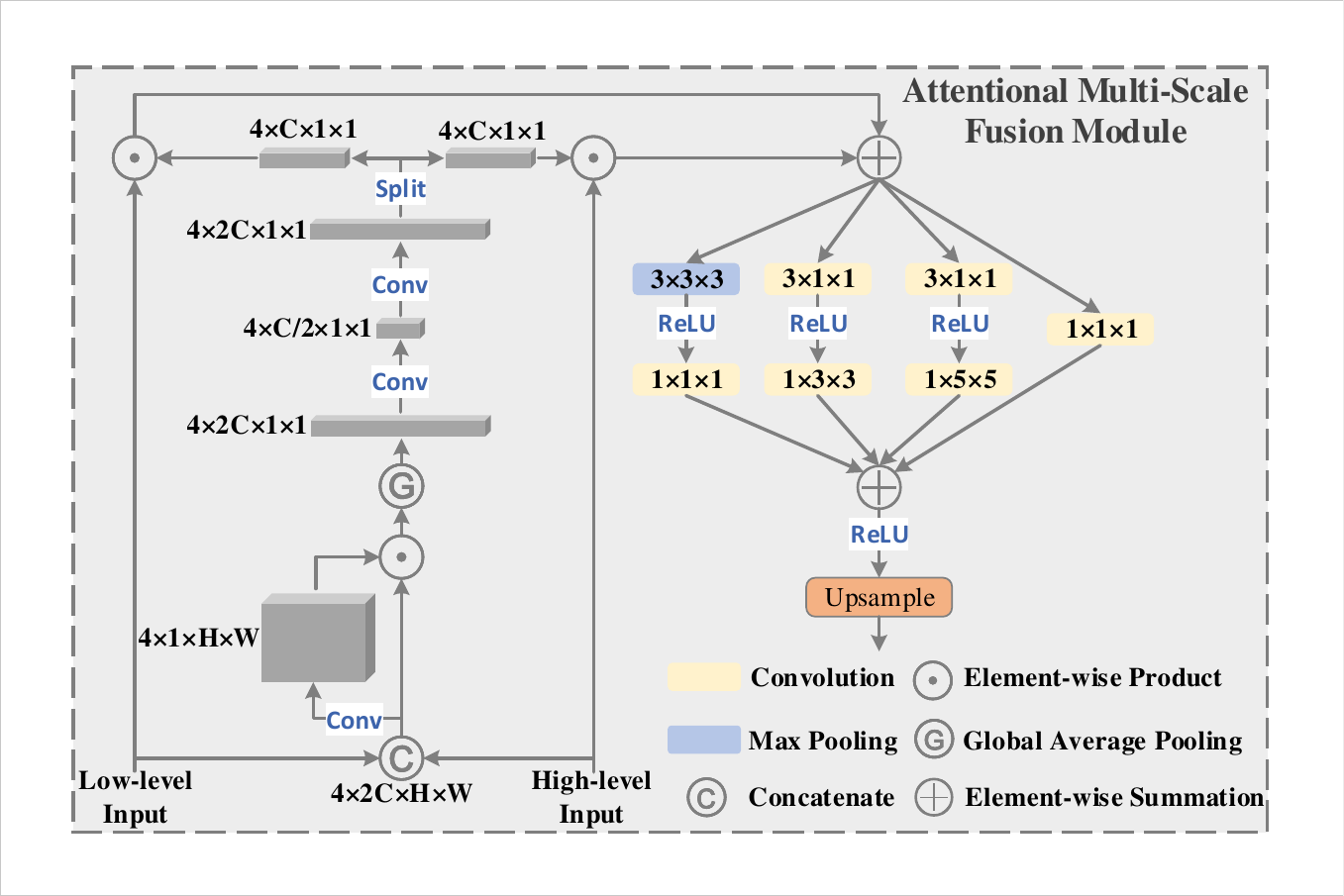}
  \caption{
      Attentional Multi-Scale Fusion (AMSF) module.
  }
  \label{Fig:AMSF}
\end{figure}
\subsection{Attentional Multi-Scale Fusion}
\label{Sec:AMSF module}
After the STSA modules, the outputs from four branches are fused for saliency inference.
The general method is the top-down feature fusion, in which the low-resolution feature is upsampled by interpolation 
and integrated with high-resolution feature using element-wise addition or concatenation.
In our model, the outputs from four STSA modules represent different contextual information in the temporal, semantic, and spatial subspaces.
Direct addition fusion is not powerful enough for this case, 
since the information gap in different subspaces is not taken into account.
To tackle this issue, we design an Attentional Multi-Scale Fusion (AMSF) module.
As depicted in \autoref{Fig:AMSF}, this module can be divided into left and right parts.
The attentional weighting operation deals with semantic gap,
and the spatio-temporal multi-scale structure alleviates information gap in spatial and temporal subspaces.

We define the output of a pair of adjacent branches as $\{\mathbf{F}_{h}, \mathbf{F}_{l}\} \in \mathbb{R}^{C \times 4 \times H \times W}$, 
representing high- and low-level features.
Concretely, in the AMSF module, $\mathbf{F}_{h}$ and $\mathbf{F}_{l}$ are first concatenated along the semantic channel, 
and masked by an attention multidimensional matrix $\mathbf{W}_{m}\in \mathbb{R}^{1 \times 4 \times H \times W}$
that is generated from $[\mathbf{F}_{h}, \mathbf{F}_{l}]$ 
to enhance the important position and weaken the invalid position, which can be formulated as:
\begin{equation}
  \mathbf{F}_{M} = \mathbf{W}_{m} \odot [\mathbf{F}_{h}, \mathbf{F}_{l}] = \mathit{\sigma}(\mathrm{Conv}([\mathbf{F}_{h}, \mathbf{F}_{l}])) \odot [\mathbf{F}_{h}, \mathbf{F}_{l}] 
\end{equation}
where $\sigma$ indicates the sigmoid activation function, $\mathrm{Conv}(\cdot)$ indicates the 3D convolutional layer,
and $\odot$ indicates the element-wise product.
After that, inspired by the SE Block \cite{hu2020SENet}, we employ a global average pooling to squeeze the spatial subspace, 
and then obtain a semantic weight matrix $[\mathbf{W}_h,\mathbf{W}_l]\in \mathbb{R}^{C \times 4 \times 1 \times 1}$
with two convolutional layers as follows:
\begin{equation}
  [\mathbf{W}_h,\mathbf{W}_l] = \mathit{\sigma}(\mathrm{Conv}(\mathit{Relu}(\mathrm{Norm}(\mathrm{Conv}(\mathrm{GAP}(\mathbf{F}_{M}))))),
\end{equation}
where $\mathit{Relu}(\cdot)$ is the rectified linear unit activation function, and $\mathrm{GAP}(\cdot)$ is the global average pooling.
Finally, the weight matrix is split to two parts,
and they are used to recalibrate $\mathbf{F}_{h}$ and $\mathbf{F}_{l}$ as folows:
\begin{equation}
  \mathbf{F}_{O} = \mathbf{F}_{h} \odot \mathbf{W}_h \oplus \mathbf{F}_{l} \odot \mathbf{W}_l,
\end{equation}
With this attention mechanism, $\mathbf{F}_{h}$ and $\mathbf{F}_{l}$ are selected in the semantic subspace
with the perception of semantic relationship between the features from adjacent branches.

In addition, as shown in the right part of \autoref{Fig:AMSF}, for spatial and temporal subspaces, we design a spatio-temporal multi-scale structure, including three convolution branches and a pooling branch, 
after the attentional weighting operation in the AMSF module.
Parallel convolutional layers with different spatial kernel sizes achieve adaptability to different spatial scales of features.
Besides, the 3D convolution and the 3D pooling 
provide the perception of context in the temporal subspace for the fusion module.

To sum up, in the AMSF module, specific calculations with trainable parameters are deployed in three subspaces,
to obtain the capability of perceiving and alleviating the spatio-temporal and semantic gaps for accurate saliency results.

\subsection{Implementation Details}
\label{Sec:implementation details}
\subsubsection{Spatio-Temporal Self-Attention Module}
The output features, from $Conv\_Block\_1$, $Conv\_Block\_2$, $Conv\_Block\_3$, and $Conv\_Block\_4$ of backbone,
have the temporal channel dimensions of $16$, $16$, $8$, and $4$, respectively.
In order to reduce memory occupancy of STSA modules, 
the temporal channel dimensions of output features from $Conv\_Block\_1$, $Conv\_Block\_2$, and $Conv\_Block\_3$ 
are uniformly compressed to $4$ by 3D convolutional layers, which are set as $4 \times 1 \times 1$ with temporal stride 4,
$4 \times 1 \times 1$ with temporal stride 4, and $2 \times 1 \times 1$ with temporal stride 2, respectively.
In the STSA module, as given in \autoref{Fig:attention}, 
for query and key embedding layers, we use a $1 \times 1 \times 1$ convolutional layer to compress the dimension of semantic channel,
and a following $1 \times 3 \times 3$ convolutional layer for spatial information.
For the value embedding layers, we adopt the asymmetric convolution (\textit{i.e.,} a $ 1 \times 3 \times 1$ convolutional layer followed by 
a $1 \times 1 \times 3$ convolutional layer) to obtain spatial information without changing the dimension of semantic channel.
These settings make the embedding layers of the STSA module cost-effective and able to capture spatial information.

Specially, in the STSA module on the branch from $Conv\_Block\_1$,
we devise a spatial bottleneck structure (\textit{i.e.,} a bottleneck structure established on the spatial subspace) 
for the STSA layers,
because the output from $Conv\_Block\_1$ has a large resolution, 
which consumes a lot of memory during dot-product attention calculation.
As shown in \autoref{Fig:bottleneck}, the spatial size of input feature is first reduced by half using a $1 \times 2 \times 2$ pooling
layer, and the pooling indices are temporarily stored.
After splitting, embedding, dot-product attention, and concatenation, the feature is restored to the initial
resolution using an unpooling layer with the pooling indices.
By this means, the dot-product attention calculation is implemented in a bottleneck of small resolution,
which greatly reduces the occupancy of memory.

\subsubsection{Attentional Multi-Scale Fusion}
In the attentional weighting operation, the spatial attention part generates the attention matrix 
$\mathbf{W}_{m}$ by: $\mathrm{Conv}(1 \times 1 \times 1) \rightarrow Sigmoid$.
After that, the semantic channel attention part gets the weight matrix $[\mathbf{W}_h,\mathbf{W}_l]$ by:
$\mathrm{GAP} \rightarrow \mathrm{Conv}(1 \times 1 \times 1) \rightarrow Relu \rightarrow \mathrm{Norm} \rightarrow \mathrm{Conv}(1 \times 1 \times 1) \rightarrow Sigmoid$.
The spatio-temporal multi-scale structure is inception-like, and the detailed settings can be clearly found 
in the right part of \autoref{Fig:AMSF}.

\subsection{Supervision and Loss Function}
\label{Sec:supervision and loss function}
\subsubsection{Supervision}
The proposed model takes successive 32 frames from a video as input, and produces a saliency map.
That is, our model predicts results for videos in a window-sliding manner.
Due to the symmetry of our model along the temporal channel, 
we supervise model training with ground truth of middle frame in the video clip, \textit{i.e.,} the 16-$\mathit{th}$ frame of 32 frames.
Therefore, to generate saliency maps for the first 15 frames and the last 16 frames of a video, 
we repeat the first frame and the last frame, respectively, to construct complete input clips.

\begin{align}
  \mathcal{L}(\mathbf{S},\mathbf{G}) = \mathrm{KL}(\mathbf{S},\mathbf{G}) - \mathrm{CC}(\mathbf{S},\mathbf{G}),
\end{align}
where $\mathbf{S}$ and $\mathbf{G}$ are the predicted saliency map and the ground truth, respectively.

\begin{table*}[t]\normalsize
  \setlength{\tabcolsep}{5pt}
  \centering
  \caption{
    Comparison Results on the Test Sets of DHF1K, Hollywood-2 and UCF Datasets.
    The Best Two are Marked by Red and Blue, Respectively.
  }
  \label{Table:DHF1K_Hollywood_UCF}
  \renewcommand{\tabcolsep}{1mm}
   \resizebox{1\textwidth}{!}{
  \begin{tabular}{|c|c|c|c|c||c|c|c|c|c||c|c|c|c||c|c|c|c|}
  \hline
  \multirow{2}{*}{Model}                      & \multirow{2}{*}{Input Size} & FLOPs                & Model Size              & Runtime         & \multicolumn{5}{c||}{DHF1K}                                                                                                                              & \multicolumn{4}{c||}{Hollywood-2}                                                                                          & \multicolumn{4}{c|}{UCF}                                                                                                   \\ \cline{6-18}
                                              &                             & (G)                    & (MB)                     & (s)              & CC $\uparrow$                & NSS $\uparrow$               & SIM $\uparrow$                & AUC $\uparrow$              & sAUC $\uparrow$                & CC $\uparrow$              & NSS  $\uparrow$              & SIM $\uparrow$               & AUC   $\uparrow$              & CC $\uparrow$               & NSS  $\uparrow$              & SIM  $\uparrow$              & AUC   $\uparrow$               \\ \hline\hline
  DeepVS \cite{jiang2018DeepVS}               & $448 \times 448$            & 819.5                & 344                       & 0.050                    & 0.344                        & 1.911                        & 0.256                        & 0.856                        & 0.583                        & 0.446                        & 2.313                        & 0.356                        & 0.887                        & 0.405                        & 2.089                        & 0.321                        & 0.870                        \\ 
  ACLNet \cite{wang2021ACLNet}                & $224 \times 224$            & 164.0                & 250                       & 0.020                    & 0.434                        & 2.354                        & 0.315                        & 0.890                        & 0.601                        & 0.623                        & 3.086                        & 0.542                        & 0.886                        & 0.510                        & 2.567                        & 0.406                        & 0.897                        \\ 
  STRA-Net \cite{lai2020STRA-Net}             & $224 \times 224$            & 540.0                & 641                       & 0.020                    & 0.458                        & 2.558                        & 0.355                        & 0.895                        & 0.663                        & 0.662                        & 3.478                        & 0.536                        & 0.923                        & 0.593                        & 3.018                        & 0.479                        & 0.910                        \\ 
  SalEMA \cite{linardos2019SalEMA}            & $256 \times 192$            & 40.0                 & 364                       & 0.010                    & 0.449                        & 2.574                        & \textbf{{\color[HTML]{FE0000} 0.466}} & 0.890                        & 0.667                        & 0.613                        & 3.186                        & 0.487                        & 0.919                        & 0.544                        & 2.638                        & 0.431                        & 0.906                        \\ 
  TASED-Net \cite{min2019TASED-Net}           & $384 \times 224$            & 91.5                 & 82                        & 0.060                    & 0.470                        & 2.667                        & 0.361                        & 0.895                        & 0.712                        & 0.646                        & 3.302                        & 0.507                        & 0.918                        & 0.582                        & 2.920                        & 0.469                        & 0.899                        \\ 
  Chen \textit{et al.}\cite{chen2021BiLSTM}   & $320 \times 256$            & 371.3                & 437                      & 0.080                    & 0.476                        & 2.685                        & 0.353                        & 0.900                        & 0.680                        & 0.661                        & 3.804                        & 0.537                        & 0.928                        & 0.599                        & 3.406                        & 0.494                        & 0.917                        \\
  SalSAC \cite{wu2020SalSAC}                  & $288 \times 160$            & -                     & 94                      & 0.020                    & 0.479                        & 2.673                        & 0.357                        & 0.896                        & 0.697                        & 0.670                        & 3.356                        & 0.529                        & 0.931                        & 0.671                        & 3.523                        & 0.534                        & \textbf{{\color[HTML]{3531FF} 0.926}} \\ 
  UNISAL \cite{droste2020UNISAL}              & $384 \times 224$            & 14.4                 & 16                       & 0.009                   & 0.490                        & 2.776                        & 0.390                        & 0.901                        & 0.691                        & 0.673                        & 3.901                        & 0.542                        & 0.934                        & 0.644                        & 3.381                        & 0.523                        & 0.918                        \\ 
  HD2S \cite{bellitto2020HD2S}                & $192 \times 128$            & -                     & 116                      & 0.030                    & 0.503                        & 2.812                        & \textbf{{\color[HTML]{3531FF} 0.406}} & 0.908                        & 0.700                        & 0.670                        & 3.352                        & 0.551                        & 0.936                        & 0.604                        & 3.114                        & 0.507                        & 0.904                        \\ 
  ViNet \cite{jain2020ViNet}                  & $384 \times 224$            & 115.0                & 124                       & 0.016                   & 0.511                        & 2.872                        & 0.381                        & 0.908                        & \textbf{{\color[HTML]{FE0000} 0.729}} & 0.693                        & 3.730                        & 0.550                        & 0.930                        & 0.673                        & 3.620                        & 0.522                        & 0.924                       \\ 
  TSFP-Net \cite{chang2021TSFP}               & $352 \times 192$            & -                     & 58                      & 0.011                  & \textbf{{\color[HTML]{3531FF} 0.517}} & \textbf{{\color[HTML]{3531FF} 2.966}} & 0.392                        & \textbf{{\color[HTML]{3531FF} 0.912}} & \textbf{{\color[HTML]{3531FF} 0.723}} & \textbf{{\color[HTML]{3531FF} 0.711}} & \textbf{{\color[HTML]{3531FF} 3.910}} & \textbf{{\color[HTML]{3531FF} 0.571}} & \textbf{{\color[HTML]{3531FF} 0.936}} & \textbf{{\color[HTML]{3531FF} 0.685}} & \textbf{{\color[HTML]{3531FF} 3.698}} & \textbf{{\color[HTML]{FE0000} 0.561}} & 0.923                        \\ \hline
  STSANet                                     & $384 \times 224$            & 193.3                & 643                       & 0.035                  & \textbf{{\color[HTML]{FE0000} 0.529}} & \textbf{{\color[HTML]{FE0000} 3.010}} & 0.383                        & \textbf{{\color[HTML]{FE0000} 0.913}} & \textbf{{\color[HTML]{3531FF} 0.723}} & \textbf{{\color[HTML]{FE0000} 0.721}} & \textbf{{\color[HTML]{FE0000} 3.927}} & \textbf{{\color[HTML]{FE0000} 0.579}} & \textbf{{\color[HTML]{FE0000} 0.938}} & \textbf{{\color[HTML]{FE0000} 0.705}} & \textbf{{\color[HTML]{FE0000} 3.908}} & \textbf{{\color[HTML]{3531FF} 0.560}} & \textbf{{\color[HTML]{FE0000} 0.936}}   \\ \hline
  \end{tabular}}
\end{table*}
\begin{table}[t]\normalsize
  \setlength{\tabcolsep}{3pt}
  \centering
  \caption{
    Comparison Results on DIEM Dataset.
  }
  \label{Table:DIEM}
  \begin{tabular}{|c|c|c|c|c|c|}
  \hline
  Model                             & Test Set                               & CC    $\uparrow$                        & NSS      $\uparrow$                     & SIM      $\uparrow$                     & AUC   $\uparrow$                        \\ \hline\hline
  DeepVS \cite{jiang2018DeepVS}     &\multirow{5}{*}{\romannumeral1}         & 0.371                        & 2.235                        & 0.238                        & 0.857                        \\ 
  ACLNet \cite{wang2021ACLNet}      &                                        & 0.396                        & 2.368                        & 0.277                        & 0.881                        \\ 
  STRA-Net \cite{lai2020STRA-Net}   &                                        & 0.408                        & 2.452                        & 0.306                        & 0.870                        \\ 
  Chen \textit{et al.}\cite{chen2021BiLSTM} &                                & 0.490                        & 2.346                        & 0.396                        & 0.889                        \\
  STSANet                           &                                        & \textbf{0.625}               & \textbf{2.618}               & \textbf{0.505}               & \textbf{0.901}               \\ \hline
  ViNet \cite{jain2020ViNet}        &\multirow{3}{*}{\romannumeral2}         & 0.626                        & 2.470                        & 0.483                        & 0.898                        \\ 
  TSFP-Net \cite{chang2021TSFP}     &                                        & 0.649                        & 2.630                        & 0.529                        & 0.905                        \\ 
  STSANet                           &                                        & \textbf{0.677}               & \textbf{2.721}               & \textbf{0.541}               & \textbf{0.906}               \\ \hline
  STSANet                           &\multirow{1}{*}{\romannumeral3}         & \textbf{0.690}               & \textbf{2.787}               & \textbf{0.548}               & \textbf{0.905}               \\ \hline
  \end{tabular}
\end{table}
\subsubsection{Loss Function}
In recent years, using the combination of saliency metrics as loss function is common and effective
in static and dynamic saliency prediction models based on deep learning \cite{2018SAM, 2020SimpleNet, 2021SalED, lai2020STRA-Net, zhang2021Recurrent}.
Accordingly, for the proposed model, we investigate the most suitable metric combination experimentally,
and finalize the loss function as follows:
Kullback-Leibler Divergence (KL) is a common measure of discrepancy between two probability distributions.
Here the $\mathrm{KL}$ loss is computed as:
\begin{equation}
  \mathrm{KL}(\mathbf{S},\mathbf{G}) = \sum_i \mathbf{G}_i \mathit{log}(\epsilon + \frac{\mathbf{G}_i}{\epsilon + \mathbf{S}_i}),
\end{equation}
where $\epsilon$ is a regularization constant.

Pearson's Correlation Coefficient (CC) loss measures dependencies between two distribution maps,
which is formulated as:
\begin{equation}
  \mathrm{CC}(\mathbf{S},\mathbf{G}) = \frac{\mathit{cov}(\mathbf{S},\mathbf{G})}{\mathit{sd}(\mathbf{S}) \times \mathit{sd}(\mathbf{G})},
\end{equation}
where $\mathit{sd}$ is standard deviation, and $\mathit{cov}$ stands for covariance.

\section{Experiments and Results}
In this section, we introduce our experiments, and present their results as well as analyses.
In \autoref{Sec:dataset}, several benchmark datasets are described.
In \autoref{Sec:training procedure}, we present the training procedure of our model.
In \autoref{Sec:metric}, we briefly introduce the saliency metrics used in the evaluation.
In \autoref{Sec:comparison with SOTA}, we compare the proposed model with other state-of-the-art models on different datasets.
In \autoref{Sec:ablation}, we detailedly study the influence of main components of our model.
In \autoref{Sec:experiment on supervision}, we compare two different supervision manners on multiple datasets.
In \autoref{Sec:faliure case}, we present some failure cases and analyze the limitations of our model as well as the difficulties of VSP task.

\subsection{Datasets}
\label{Sec:dataset}
\subsubsection{DHF1K \cite{wang2021ACLNet}}
It is a large and diverse dataset, including 1K 30 fps videos with $640 \times 360$ resolution,
which are split to 600, 100, and 300 as training, validation, and testing sets, respectively.
The corresponding free-viewing data is collected from 17 observers by the eye tracker.
The ground truths of testing videos are held out for the evaluation on benchmark website\footnote{https://mmcheng.net/videosal/\label{DHF1K_web}}.

\subsubsection{Hollywood-2 \cite{hollywoodUCF}}
It contains 1,707 videos from Hollywood movies.
The annotations come from 19 observers, 3 of which are in a free-viewing mode and the others are 
driven by action and context recognition tasks.
Following the usual split, we use 823 and 884 videos as training and testing sets, respectively.

\subsubsection{UCF \cite{hollywoodUCF}}
It consists of 150 videos including kinds of sports action classes,
and the annotations are collected in a task-driven manner.
In our paper, we adopt the same split as \cite{wang2021ACLNet} with 103 video for training and 47 videos for testing.

\subsubsection{DIEM \cite{DIEM}}
It has 84 videos based on advertisements, documentaries, game trailers, and movie trailers, \textit{etc}.
The annotations of them are collected from about 50 observers in a free-viewing manner.
Following \cite{borji2013comparative, lai2020STRA-Net}, we adopt the same 20 videos as testing set,
and the rest as training set.

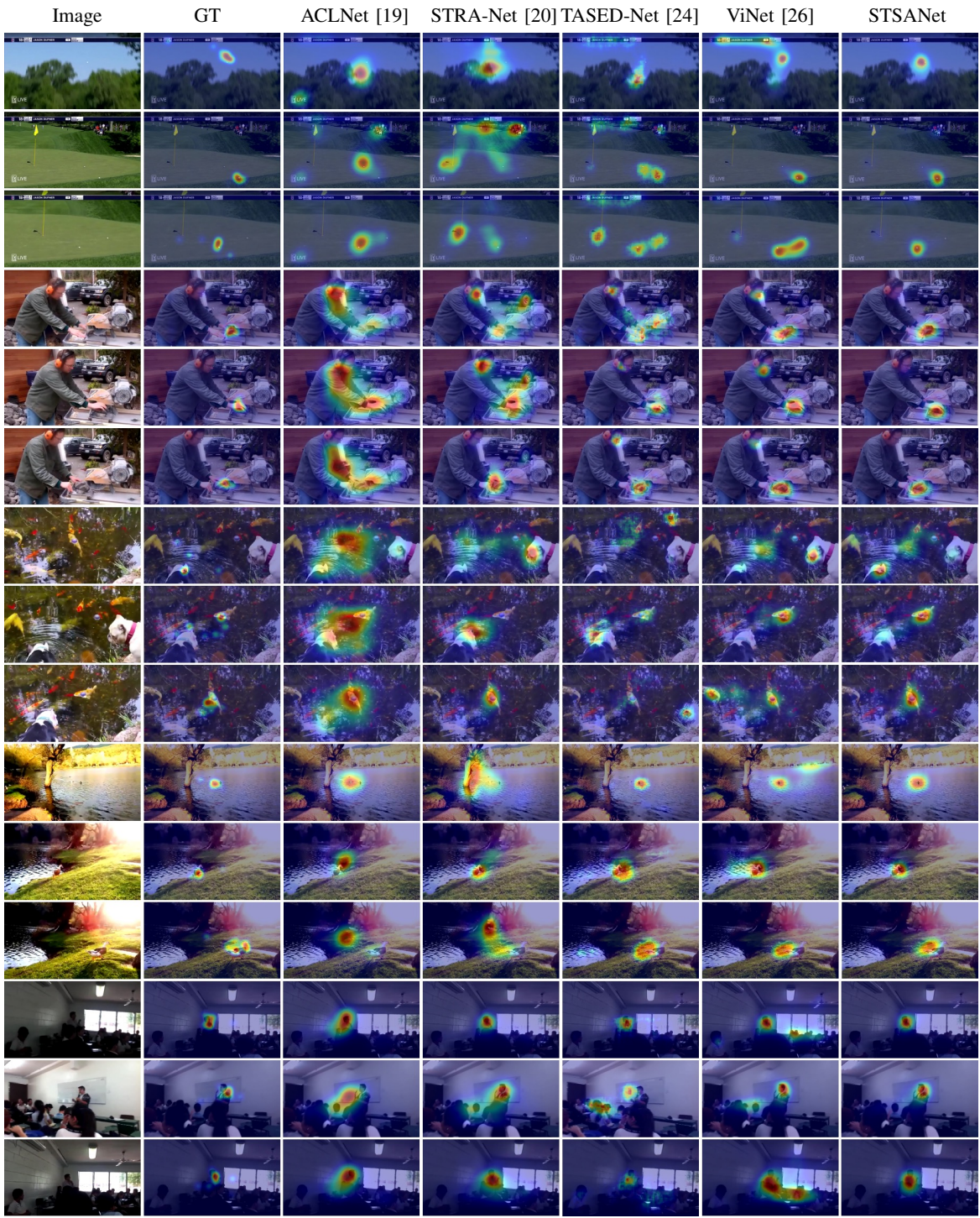
\begin{figure*}[t]  
  \centering
  \small
  \begin{overpic}[width=.85\textwidth]{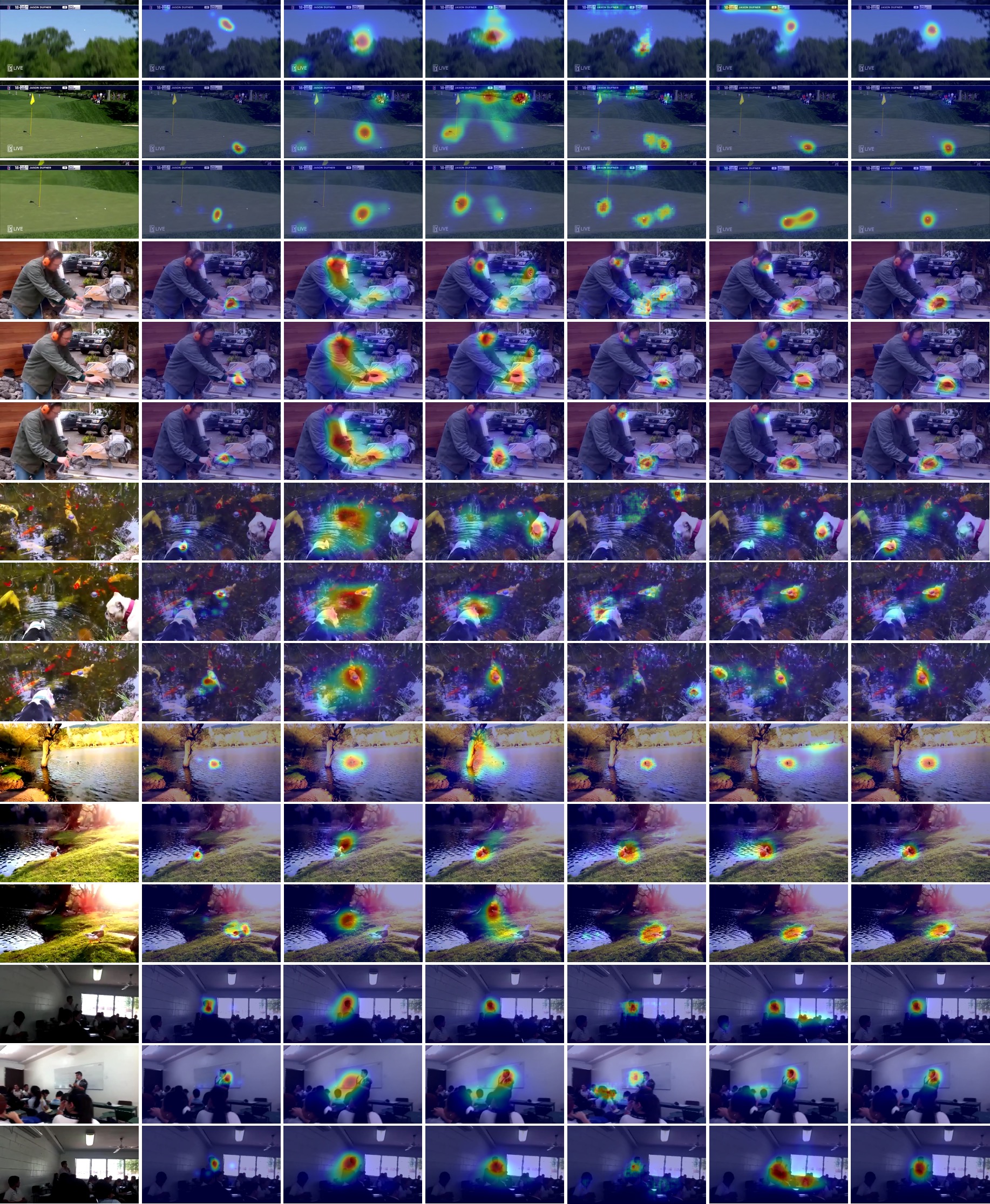}
    \put(4,101){Image}
    \put(16,101){GT}
    \put(25,101){ACLNet \cite{wang2021ACLNet}}
    \put(36,101){STRA-Net \cite{lai2020STRA-Net}}
    \put(47,101){TASED-Net \cite{min2019TASED-Net}}
    \put(61,101){ViNet \cite{jain2020ViNet}}
    \put(73,101){STSANet}
  \end{overpic}
  \caption{
    Qualitative comparisons with start-of-the-art video saliency models on various categories of videos,
    each of which samples three frames for display.
  }
  \label{Fig:results}
\end{figure*}
\subsection{Training Procedure}
\label{Sec:training procedure}
The proposed model is implemented on two  NVIDIA TITAN Xp GPUs using \emph{Pytorch} \cite{Pytorch}.
We initialize the backbone of our model with the weights from S3D\cite{xie2018S3D} pre-trained on the Kinetics dataset\cite{2017kinetics},
and the remaining network is initialized by default settings of \emph{Pytorch}.
We train the whole model with the Adam optimizer\cite{Adam}, 
and the initial learning rate is set to $10^{-4}$, which is decreased 10 times when the training loss has been saturated.

Our model is first trained with the training set of DHF1K dataset, and the validation set is used to monitor the convergence.
Then we test the results on the DHF1K benchmark.
For the Hollywood-2, UCF, and DIEM datasets, we fine-tune the proposed model from the weights trained on DHF1K,
and use the testing sets to monitor the convergence.
All input frames are resized to $384 \times 224$ and the batch size is set to 3.

\subsection{Metrics}
\label{Sec:metric}
Saliency metrics involved in our experimental results and analyses include Normalized Scanpath Saliency (NSS), 
Pearson's Correlation Coefficient (CC), Similarity (SIM), Kullback–Leibler Divergence (KL), 
and variants of Area Under ROC Curve (AUC (AUC-Judd) and sAUC (shuffled AUC)).
Specifically, the distribution-based metrics, including SIM, CC, and KL, 
are obtained by comparing results with the fixation continuous maps.
The location-based metrics, containing NSS, AUC-Judd, and sAUC, are calculated with the binary maps of fixation points.
More specific characteristics of saliency metrics can be found in \cite{bylinskii2019metrics}.

\begin{table}[t]\normalsize
  \setlength{\tabcolsep}{3.3pt}
  \centering
  \caption{
    Ablation Study on Main Components of the Proposed Model.
  }
  \label{Table:ablation}
  \begin{tabular}{|c|c|c|c|c|c|}
  \hline
  Model                                        & CC $\uparrow$  & NSS $\uparrow$ & SIM $\uparrow$ & AUC $\uparrow$ & KL $\downarrow$  \\ \hline \hline
  Single-stream                                & 0.506          & 2.839          & 0.387          & 0.910          & 1.462          \\ 
  UNet-like                    & 0.523          & 2.983          & 0.396          & 0.917          & 1.388          \\ 
  UNet-like + STSA             & 0.534          & 3.049          & 0.403          & 0.919          & 1.362          \\ 
  UNet-like + AMSF             & 0.531          & 3.033          & 0.406          & 0.919          & 1.359          \\ 
  \hline
  Ours      & \textbf{0.539} & \textbf{3.082} & \textbf{0.411} & \textbf{0.920} & \textbf{1.344} \\ \hline
  \end{tabular}
\end{table}
\subsection{Comparison with the State-of-the-Art Models}
\label{Sec:comparison with SOTA}
We compare our model with state-of-the-art VSP models based on deep learning in recent years, 
including DeepVS \cite{jiang2018DeepVS}, ACLNet \cite{wang2021ACLNet}, STRA-Net \cite{lai2020STRA-Net}, SalEMA \cite{linardos2019SalEMA}, 
Chen \textit{et al.}\cite{chen2021BiLSTM}, TASED-Net \cite{min2019TASED-Net}, SalSAC \cite{wu2020SalSAC}, UNISAL \cite{droste2020UNISAL}, 
HD2S \cite{bellitto2020HD2S}, ViNet \cite{jain2020ViNet}, and TSFP-Net \cite{chang2021TSFP},
on the DHF1K benchmark and the testing sets of Hollywood-2, UCF, and DIEM datasets.

\autoref{Table:DHF1K_Hollywood_UCF} reports a range of attributes of models, 
where input size, model size, and runtime mostly come from corresponding papers and the DHF1K benchmark website\textsuperscript{\ref{DHF1K_web}},
and FLOPs is measured using the publicly available codes of the models.
In terms of these attributes, our STSANet is at an intermediate level of computational efficiency.
In fact, the STSA module is a major contributor to computational consumption,
but it also results in considerable performance gains for the STSANet. 

On the DHF1K benchmark\textsuperscript{\ref{DHF1K_web}}, 
our model outperforms all other models on CC, NSS, and AUC metrics, and ranks second on sAUC.
More concretely, on CC and NSS, the performance of our model exceeds that of other models by $2.3\%$ and $1.5\%$, respectively. 
The SalEMA model shows the highest score on SIM metric, however,
it gets low scores on other four metrics compared with state-of-the-art models.

The results on the test sets of Hollywood-2 and UCF datasets are shown in the right of \autoref{Table:DHF1K_Hollywood_UCF},
where results of other models come from their corresponding papers and the DHF1K benchmark website\textsuperscript{\ref{DHF1K_web}}.
On the Hollywood-2 dataset, our method achieves the best performance in terms of all metrics.
On the UCF dataset, our model outperforms others by a large margin on CC and NSS metrics, and ranks first on AUC metric.
Specifically, compared with the second best model, our model has improved by $2.9\%$ and $5.7\%$ on CC and NSS, respectively.
On SIM metric, the score of our model ranks second but very close to the first one.

The results on the DIEM datasets are reported in \autoref{Table:DIEM}.
The LSTM-based models, including DeepVS \cite{jiang2018DeepVS}, ACLNet \cite{wang2021ACLNet}, STRA-Net \cite{lai2020STRA-Net},
and Chen \textit{et al.} \cite{chen2021BiLSTM}, all use the first 300 frames of 20 test videos as test set (Test set \romannumeral1) in the corresponding papers,
while the two 3D convolution-based models, ViNet \cite{jain2020ViNet} and TSFP-Net \cite{chang2021TSFP},
use 17 of 20 test videos as test set (Test set \romannumeral2).
Accordingly, we evaluate our model on different test sets for fair comparisons with other models.
On the Test set \romannumeral1, our model outperforms others by a large margin(\textit{e.g.} $\frac{0.625-0.490}{0.490} = 27.6\%$ on CC).
Besides, our model achieves a substantial improvement compared with other 3D convolution-based models on the Test set \romannumeral2,
such as $4.3\%$ (0.649 to 0.677) on CC and $3.5\%$ (2.630 to 2.721) on NSS.
Finally, we present the results of our model on all 20 test videos of DIEM dataset (Test set \romannumeral3).

\begin{figure}[t]
  \centering
  \subfigure[]{
  \label{Fig:STSA_variant_1} 
  \includegraphics[width=.49\textwidth]{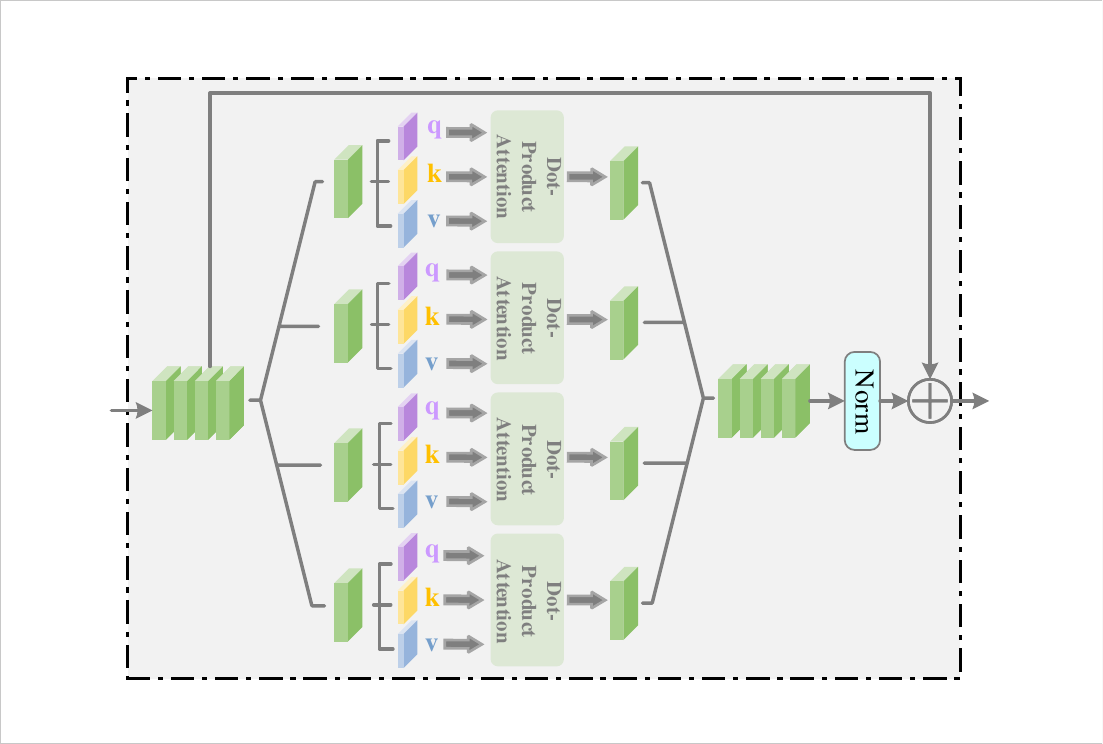}}
  \subfigure[]{
  \label{Fig:STSA_variant_2} 
  \includegraphics[width=.49\textwidth]{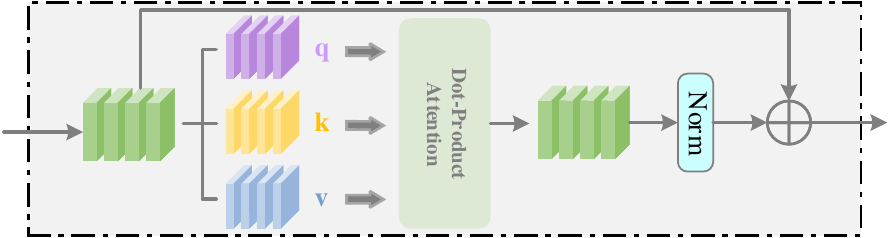}}
  \caption{
    Structures of two variant of STSA module: (a) \textit{w/o} Temporal Relations. 
    (b) Single Similarity Matrix.
  }
\end{figure}
In \autoref{Fig:results}, we further qualitatively compare our model with some representative state-of-the-art VSP models,
including LSTM-based ACLNet \cite{wang2021ACLNet} and STRA-Net \cite{lai2020STRA-Net}, 
as well as 3D convolution-based TASED-Net \cite{min2019TASED-Net} and ViNet \cite{jain2020ViNet}.
It can be clearly observed that our model achieves more accurate results than others on different indoor and outdoor videos.

\subsection{Ablation Analysis}
\label{Sec:ablation}
In this section, we conduct comprehensive ablation experiments for STSANet on the validation set of DHF1K dataset,
which contains around 60K frames. 
We first investigate the contribution of main components in our model,
and then further provide detailed ablation studies on our STSA and AMSF modules, respectively.

\subsubsection{The Contributions of Main Components}
We quantitatively evaluate the effectiveness of main components in the proposed model,
and the results are reported in \autoref{Table:ablation}.
We construct a single-stream network based on the 3D backbone as a baseline,
in which feature maps are gradually restored to initial image size by 3D convolution and trilinear interpolation layers after backbone.
Next, the UNet-like structure is added over the baseline for exploiting the multi-level information of encoder,
which brings obvious improvements in terms of all metrics.
Based on the UNet-like encoder-decoder architecture, the scores of CC and NSS increase $2.1\%$ (0.523 to 0.534) 
and $2.2\%$ (2.983 to 3.049), respectively, after adding the proposed STSA modules.
As for the AMSF module, replacing the traditional top-down feature fusion with AMSF improves the CC score by $1.5\%$.
Besides, adding the combination of STSA and AMSF over the UNet-like encoder-decoder model increases the scores of CC, NSS, and SIM by $3.1\%$ (0.523 to 0.539),
$3.3\%$ (2.983 to 3.082), and $3.8\%$ (0.396 to 0.411), respectively, and optimizes the score of KL by $3.2\%$ (1.388 to 1.344).
Overall, continuous performance improvements are shown when adding main components into the baseline in turn.
From the baseline to the full settings, the scores of CC, NSS, and KL are totally optimized by 
$6.5\%$ (0.506 to 0.539), $8.6\%$ (2.839 to 3.082), and $8.1\%$ (1.462 to 1.344), respectively.

\begin{table}[t]\normalsize
  \setlength{\tabcolsep}{1.8pt}
  \centering
  \caption{
    The Upper Part is Ablation Study on the Structure of STSA Module,
    where \textit{w/o} Temporal Relations Means Implementing Self-Attention Mechanism Separately on Each Temporal Channel,
    and Single Similarity Matrix Means Implementing Self-Attention Mechanism Directly on the Entire Feature.
    The Lower Part is Ablation Study on the STSA Modules at Different Levels,
    where \textit{rm} STSA\_n Means Removing the STSA Module after $Conv\_Block\_n$.
  }
  \label{Table:STSA}
  \begin{tabular}{|c|c|c|c|c|c|}
  \hline
  Model                                                 & CC $\uparrow$  & NSS $\uparrow$ & SIM $\uparrow$ & AUC $\uparrow$ & KL $\downarrow$\\ \hline\hline
  \textit{w/o} Temporal Relations                                & 0.532          & 3.040          & 0.402          & 0.919          & 1.358          \\
  Single Similarity Matrix                                       & 0.532          & 3.021          & 0.395          & 0.919          & 1.357          \\
  \textit{w/o} STSA Layer \uppercase\expandafter{\romannumeral1} & 0.530          & 3.026          & 0.397          & 0.918          & 1.370          \\ 
  \textit{w/o} STSA Layer \uppercase\expandafter{\romannumeral2} & 0.533          & 3.026          & 0.404          & 0.919          & 1.357          \\ 
  \hline
  \textit{rm} STSA\_1                                            & 0.534          & 3.051          & 0.404          & 0.919          & 1.353          \\ 
  \textit{rm} STSA\_2                                            & 0.532          & 3.022          & 0.401          & 0.919          & 1.356          \\ 
  \textit{rm} STSA\_3                                            & 0.535          & 3.056          & 0.407          & 0.919          & 1.350          \\ 
  \textit{rm} STSA\_4                                            & 0.536          & 3.047          & 0.405          & 0.919          & 1.352          \\ 
  \hline
  Ours                                                  & \textbf{0.539} & \textbf{3.082} & \textbf{0.411} & \textbf{0.920} & \textbf{1.344} \\ \hline
  \end{tabular}
\end{table}
\subsubsection{Ablation Study on STSA Module}
\label{Sec:ablation on STSA}
As reported in \autoref{Table:STSA}, we conduct an in-depth ablation study on our STSA module.
As mentioned in \autoref{Sec:STSA module}, the STSA module achieves that each feature at different time steps is 
directly updated by long-range spatio-temporal relations with the others through the cooperation of two STSA layers,
for enhancing the relational features between time steps.
To study the importance of temporal relations contained in two STSA layers, we construct a variant of STSA module without 
capturing temporal relations between time steps, namely \textit{w/o} Temporal Relations, as depicted in \autoref{Fig:STSA_variant_1}.
In this variant, self-attention mechanism is implemented separately at different time steps.
Specifically, for feature at each time step, its own query and key are multiplied for similarity matrix 
that is then integrated to value by matrix multiplication.
Consequently, this variant captures long-distance spatial relations separately at each time step, 
without temporal relations between different time steps.
We replace the STSA module with this variant in our model and the evaluation results indicate 
the contribution of capturing long-range relations between spatio-temporal features of different time steps in our STSA module.

In the second variant named as Single Similarity Matrix, as depicted in \autoref{Fig:STSA_variant_2},
self-attention mechanism is applied to the whole feature $\mathbf{F} = 
[\mathbf{F}^{1}, \mathbf{F}^{2}, \mathbf{F}^{3}, \mathbf{F}^{4}] \in \mathbb{R}^{C \times 4 \times H \times W}$. 
Concretely, the input feature $\mathbf{F}$ is transformed to query $\mathbf{F}^{t}_q \in \mathbb{R}^{C/2 \times 4 \times H \times W}$,
key $\mathbf{F}^{t}_k \in \mathbb{R}^{C/2 \times 4 \times H \times W}$, and value $\mathbf{F}^{t}_v \in \mathbb{R}^{C \times 4 \times H \times W}$
by embedding layers.
In the dot-product attention, a single similarity matrix with size $4HW \times 4HW$ is generated from query and key,
to capture spatio-temporal relationship.
The experimental results of the Single Similarity Matrix show worse performance than the proposed model as reported in \autoref{Table:STSA}. 
In this variant, the elements in the similarity matrix are selected in temporal subspace during the softmax function.
For example, if the elements representing the relation between $\mathbf{F}^{1}$ and itself get high values, 
the elements representing the relations between $\mathbf{F}^{1}$ and sub-features at other time steps ($\mathbf{F}^{2}$, $\mathbf{F}^{3}$, 
and $\mathbf{F}^{4}$) will get low values,
which amounts to a disguised weakening of these elements,
and hence leads to an inability to adequately capture the relations between $\mathbf{F}^{1}$ and sub-features at other time steps.
Our STSA module avoids this drawback,
because it directly captures long-range relations between each sub-feature and 
the other three in two STSA layers.

\begin{figure}[t]  
  \centering
  \includegraphics[width=.49\textwidth]{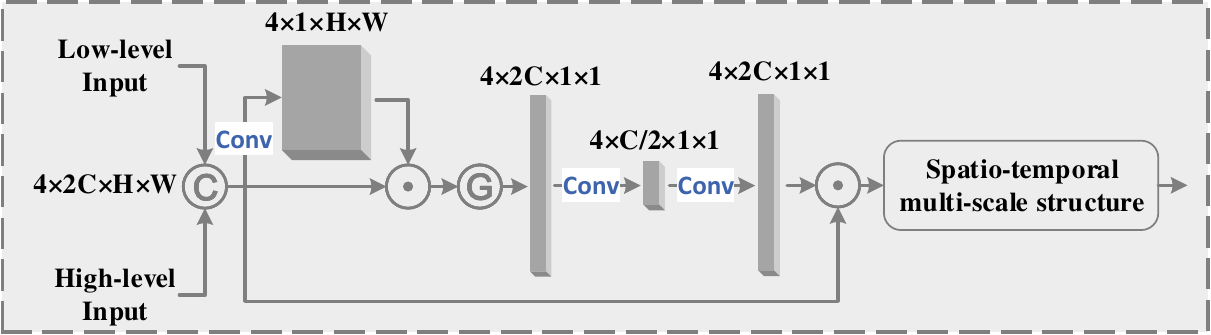}
  \caption{
      Structure of the variant of AMSF module: replacing addition fusion with concatenation fusion in AMSF module.
  }
  \label{Fig:AMSF_variant}
\end{figure}
\begin{table}[t]\normalsize
  \setlength{\tabcolsep}{5.2pt}
  \centering
  \caption{
    Ablation Study on the AMSF Module.
    Concat-Based Means Replacing Addition Fusion with Concatenation Fusion.
    AM and STMS Indicate the Attention Weighting Operation and Spatio-Temporal Multi-Scale Structure, respectively.
  }
  \label{Table:AMSF}
  \begin{tabular}{|c|c|c|c|c|c|}
  \hline
  Model                                          & CC $\uparrow$  & NSS $\uparrow$ & SIM $\uparrow$ & AUC $\uparrow$ & KL $\downarrow$\\ \hline\hline
  Concat-Based                                   & 0.532          & 3.015         & 0.397           & 0.919          & 1.355          \\ 
  \textit{w/o} AW                                & 0.533          & 3.033         & 0.401           & 0.919          & 1.352          \\ 
  \textit{w/o} STMS                              & 0.528          & 3.019         & 0.403           & 0.918          & 1.372          \\ \hline
  Ours                                          & \textbf{0.539} & \textbf{3.082} & \textbf{0.411}  & \textbf{0.920} & \textbf{1.344} \\ \hline
  \end{tabular}
\end{table}
Furthermore, in order to inspect the necessity of two STSA layers in the STSA module,
the STSA layer \uppercase\expandafter{\romannumeral1} and STSA layer \uppercase\expandafter{\romannumeral2} are removed separately.
The comparison results demonstrate that removing either STSA layer \uppercase\expandafter{\romannumeral1} or STSA layer \uppercase\expandafter{\romannumeral2}
results in degradation of performance compared with the full settings,
in which each of features at different time steps can be updated by long-range relations with spatio-temporal features of all the other time steps through the cooperation of two STSA layers.

Moreover, to validate the effectiveness of STSA modules employed on different levels,
we separately remove one of four STSA modules from the proposed STSANet for evaluation.
As shown in the lower part of \autoref{Table:STSA}, the performance deteriorates in terms of all metrics after removing STSA module at any level,
which suggests that all STSA modules at different levels have contribution to saliency results.

\subsubsection{Ablation Study on AMSF Module}
To further verify the contribution of the design of AMSF module,
we change the AMSF module by removing the attentional weighting (AW) operation and spatio-temporal multi-scale (STMS) structure, respectively.
As shown in \autoref{Table:AMSF}, the two variants with incomplete structures result in performance degradation in terms of all metrics.
Our complete AMSF modules integrates multi-level features with the perception of context in all three subspaces for better saliency results.

Besides, as shown in \autoref{Fig:AMSF_variant}, we construct a variant of the AMSF module about feature fusion, namely Concat-Based,
which replaces addition fusion with concatenation fusion in the AMSF module.
The experimental comparison shows that the AMSF module helps the proposed model go to a better convergence 
compared with the Concat-Based variant.
In our AMSF module, addition fusion allows highlighting the intersecting saliency regions in high- and low-level feature maps,
which helps to collectively consider the multi-level saliency information during feature fusion.
Moreover, addition fusion brings fewer parameters than concatenation fusion.

\begin{table}[t]
  \setlength{\tabcolsep}{3pt}
  \centering
  \caption{
    Comparison Results of Different Supervision Manners on DHF1K, Hollywood-2, and UCF Datasets.
  }
  \label{Table:supervision}
  \begin{tabular}{|c|c|c|c|c|c|c|}
  \hline
  Dataset                      & Supervision       & CC $\uparrow$  & NSS $\uparrow$ & SIM $\uparrow$ & AUC $\uparrow$ & KL $\downarrow$\\ 
  \hline \hline
  \multirow{2}{*}{DHF1K}       & middle-supervised & \textbf{0.539} & \textbf{3.082} & \textbf{0.411} & \textbf{0.920} & \textbf{1.344} \\  
                               & last-supervised   & 0.537          & 3.057          & 0.405          & \textbf{0.920} & \textbf{1.344} \\ \hline
  \multirow{2}{*}{Hollywood-2} & middle-supervised & 0.721          & \textbf{3.927} & \textbf{0.579} & 0.938          & 0.769          \\ 
                               & last-supervised   & \textbf{0.722} & 3.908          & 0.576          & \textbf{0.939} & \textbf{0.764} \\ \hline
  \multirow{2}{*}{UCF}         & middle-supervised & \textbf{0.705} & \textbf{3.908} & \textbf{0.560} & \textbf{0.936} & \textbf{0.851} \\ 
                               & last-supervised   & 0.700          & 3.789          & 0.541          & 0.935          & 0.873          \\ \hline
  \end{tabular}
\end{table}
\subsection{Discussion on Supervision Manner}
\label{Sec:experiment on supervision}
Previous work \cite{min2019TASED-Net} used the ground truths corresponding to last frames of input video clips for supervise learning.
In our work, as mentioned in \autoref{Sec:supervision and loss function}, 
the ground truth of middle frame in the video clip is used to supervise model training.
We refer to the two supervision manners as middle-supervised and last-supervised for short, and compare them experimentally.
As show in \autoref{Table:supervision}, we adopt two different supervision manners to train our model on DHF1K, Hollywood-2, and UCF datasets, respectively.
The evaluation results show that the middle-supervised and last-supervised have close performance on the DHF1K and Hollywood-2 datasets.
On the UCF dataset, the middle-supervised manner presents a relative advantage in terms of all metrics.
As a result, other experiments are implemented in the middle-supervised manner.

\subsection{Failure Cases and Analyses}
\label{Sec:faliure case}
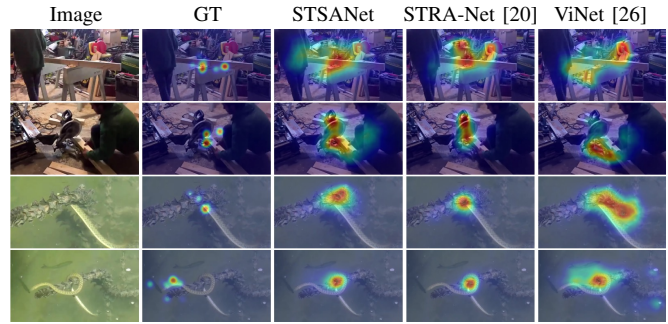
\begin{figure}[t]  
  \centering
  \footnotesize
  \begin{overpic}[width=.48\textwidth]{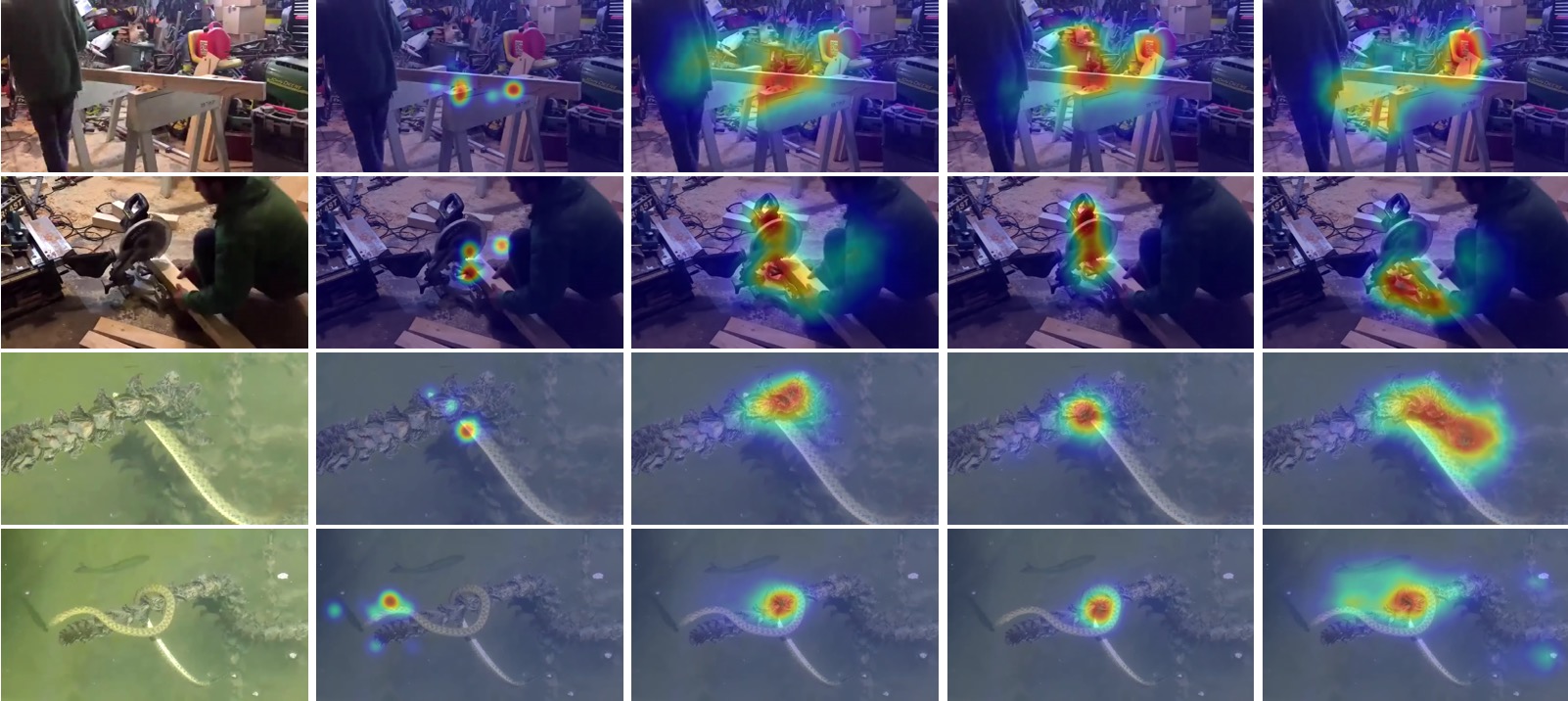}
    \put(6,46){Image}
    \put(28,46){GT}
    \put(43,46){STSANet}
    \put(60,46){STRA-Net \cite{lai2020STRA-Net}}
    \put(83,46){ViNet \cite{jain2020ViNet}}
  \end{overpic}
  \caption{
        Some failure cases on DHF1K dataset.
  }
  \label{Fig:failure}
\end{figure}

Here we present some failure cases and analyze the limitations of our model as well as the difficulties of VSP task.
In \autoref{Fig:failure}, for the first video (the first two rows), human fixations are mainly on the wood being processed. 
However, the results of computational models are scattered on other objects.
Varied scene and all sorts of objects make it difficult for models to accurately localize the wood, 
\textit{i.e.}, the main character of the video.
On the other hand, our STSANet infers saliency results only from a video clip, 
so the global context from an entire video is not taken into account, 
which causes difficulty in highlighting the main character in such a complex video.
In the second video (the last two rows), human fixations are distributed on the interaction of the water snake with other objects,
\textit{i.e.}, the water snake and water plants (the third row), and the water snake and fish (the last row).
Although models are able to capture the motions of the snake, it is hard for them to perceive the interaction, 
and thus they fail to precisely locate the part of the snake's body.

\section{Conclusion}
This paper proposes a novel Spatio-Temporal Self-Attention 3D neural network (STSANet) for video saliency prediction.
We employ Spatio-Temporal Self-Attention (STSA) modules at different levels of the 3D convolutional backbone to overcome the locality of 3D convolution.
In each STSA module, spatio-temporal features are split along the temporal channel and features at different time steps are transformed by embedding layers,
then dot-product attention calculation is implemented between features of different time steps.
By this means, at multiple levels, features at each time step can be updated by long-range dependencies with features of other time steps.
The integration of 3D networks and spatio-temporal self-attention mechanism brings performance gains as shown in ablation experiments.
Accordingly, this method has the potential to be applied to other video tasks.
Furthermore, we design an Attentional Multi-Scale Fusion (AMSF) module for the integration of multi-level spatio-temporal features.
The AMSF module contains an attentional weighting operation and a spatio-temporal multi-scale structure,
which separately aim to alleviate the semantic and spatio-temporal gaps between features of different levels.
Extensive experiments demonstrate outstanding performance of the proposed model compared with all state-of-the-art video saliency prediction methods.


\ifCLASSOPTIONcaptionsoff
  \newpage
\fi



\bibliographystyle{IEEEtran}
\bibliography{STSANet}




\end{document}